\documentclass[journal]{IEEEtran}
\usepackage{amsmath,amsfonts}
\usepackage{algorithmic}
\usepackage{algorithm}
\usepackage{array}
\usepackage{subcaption}
\usepackage{textcomp}
\usepackage{stfloats}
\usepackage{url}
\usepackage{verbatim}
\usepackage{graphicx}
\usepackage{cite}
\usepackage{bbding} 
\usepackage{booktabs}
\usepackage{xcolor}
\usepackage{mathtools}
\usepackage{xspace}
\usepackage[pagebackref,breaklinks,colorlinks]{hyperref}
\usepackage[abs]{overpic}
\usepackage{multirow}
\usepackage{amssymb}
\usepackage{pgfplots}
\usepackage{picins}

\usepackage{hyperref}
\hypersetup{  
	colorlinks = true,     % 更改链接颜色  
	linkcolor = blue,      % 链接颜色  
	urlcolor = blue,       % URL 颜色  
	citecolor = blue,      % 引用颜色    
	linkbordercolor = red
}

\usepackage{cleveref}

\usepackage{pgfplots} 

\makeatletter
\DeclareRobustCommand\onedot{\futurelet\@let@token\@onedot}
\def\@onedot{\ifx\@let@token.\else.\null\fi\xspace}
\def\eg{\emph{e.g}\onedot} 
\def\ie{\emph{i.e}\onedot}

\makeatother

\begin{document}

\title{Learning Dual-Camera Smooth Zoom via \\ Decoupled Geometric-Photometric Data Synthesis}    

\author{Junjie Fan, Zhilu Zhang, Renlong Wu, Dazhi Zhang and Wangmeng Zuo,~\IEEEmembership{Senior Member,~IEEE}% <-this % stops a space
\IEEEcompsocitemizethanks{\IEEEcompsocthanksitem Junjie Fan and Dazhi Zhang are with Computational Mathematics, Harbin Institute of Technology, Harbin, China. (E-mail: 25B912036@stu.hit.edu.cn; zhangdazhi@hit.edu.cn) Zhilu Zhang, Renlong Wu, and Wangmeng Zuo are with the Faculty of Computing, Harbin Institute of Technology, Harbin, China. (E-mail: cszlzhang@outlook.com; hirenlongwu@gmail.com; cswmzuo@gmail.com)
}% <-this % stops an unwanted space
% \thanks{Manuscript received xxx; revised xxx.} \protect\\ 
% \thanks{Manuscript received April 19, 2005; revised August 26, 2015.}
}

% The paper headers
% \markboth{Journal of \LaTeX\ Class Files,~Vol.~14, No.~8, August~2021}%
% {Shell \MakeLowercase{\textit{et al.}}: A Sample Article Using IEEEtran.cls for IEEE Journals}

\maketitle

\begin{abstract}
     When zooming between dual cameras on smartphones, prominent discontinuities in both geometric perspective and color rendition inevitably degrade the visual experience of users. To address this issue, we introduce the dual-camera smooth zoom (DCSZ) task, aiming to synthesize continuous intermediate frames for fluid zooming.  However, directly applying existing video frame interpolation (VFI) models to this task is severely hindered by substantial domain gaps and the practical intractability of capturing real-world continuous ground-truth. To this end, we propose a novel data synthesis framework to generate high-fidelity DCSZ training sequences, named ZoomGS v2. It explicitly decouples geometric and color transformations from 3D scene representation modeling, thereby enabling extensive data generation from easily accessible single-camera 3D scenes. Furthermore, we develop ZoomFI, a specialized VFI network tailored to the DCSZ task, which leverages complementary bidirectional warping for comprehensive dual-camera information fusion. Extensive experiments on test datasets captured using two mobile phones demonstrate that training with the constructed DCSZ data significantly improves the performance of various VFI models. Moreover, ZoomFI achieves state-of-the-art results in both quantitative metrics and visual quality. The datasets, code, and pretrained models are available at \url{https://hitjunjief.github.io/DCSZ-v2}.
\end{abstract}


\section{Introduction}
\begin{figure*}[t!]
	\centering
	\includegraphics[width=1.0\linewidth]{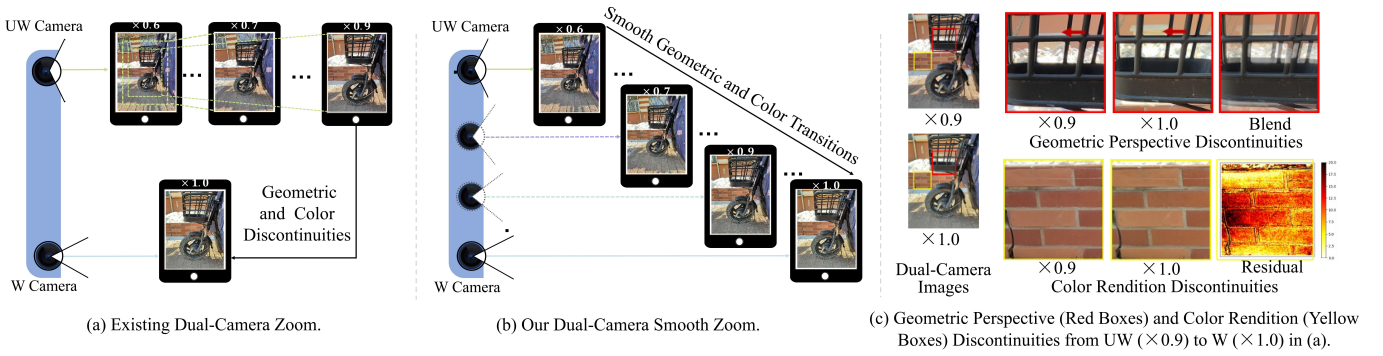}
	\caption{
		(a) Existing dual-camera digital zoom.
		For zoom transitions between the UW and W cameras (\ie, from $\times 0.6$ to $\times 1.0$), commercial smartphones generally crop the corresponding region from the UW image and resize it to the target dimensions.
		When the zoom factor transitions from $\times 0.9$ to $\times 1.0$, the active lens switches abruptly from UW to W, inducing severe geometric perspective and color rendition discontinuities in the preview.
		(b) The proposed dual-camera smooth zoom (DCSZ) pipeline.
		(c) Visual illustration of geometric and color rendition discontinuities in conventional dual-camera zoom previews.
	}
	\label{fig:intro}
    \vspace{0mm}
\end{figure*}

\IEEEPARstart{D}{ue} to physical space constraints, modern smartphones predominantly deploy multiple fixed-focal-length cameras rather than variable-focal-length lenses for digital zoom~\cite{blahnik2021smartphone}.
In particular, dual-camera configurations are commonly deployed,  typically pairing a wide-angle (W, \eg, $\times 1.0$) lens for primary imaging with an ultra-wide-angle (UW, \eg, $\times 0.5$ or $\times 0.6$) lens to provide an expanded field of view.
For zooming between the UW and W cameras, commercial smartphones (\eg, manufactured by Xiaomi, Huawei, Oppo, and Vivo) conventionally crop the central region of the UW image and digitally scale it up to the target resolution, as illustrated in Fig.~\ref{fig:intro}(a).
However, once the zoom magnification reaches the transition threshold (\eg, switching from $\times 0.9$ to $\times 1.0$), the system abruptly switches from the UW camera to the W camera.
Due to the distinct deployment locations and internal characteristics~\cite{zhang2022self,alzayer2023dc2,abdelhamed2021leveraging} of the two sensors, this switch induces prominent preview discontinuities, manifested as abrupt shifts in both geometric perspective and color rendition (see Fig.~\ref{fig:intro}(c)).
% 
% It substantially degrades the visual experience of users.
These discontinuities substantially degrade the user experience.
To overcome this limitation, the task of Dual-Camera Smooth Zoom (DCSZ) has emerged, which aims to synthesize continuous frames between dual cameras to deliver a fluid zoom preview on mobile phones, as shown in Fig.~\ref{fig:intro}(b).

Video frame interpolation (VFI) techniques~\cite{AMT,VFIFormer,wu2024perception,guo2024generalizable,zhong2024clearer} offer a potential remedy by synthesizing intermediate frames between the endpoint views of dual cameras.
However, directly applying off-the-shelf VFI models often fails to produce satisfactory results.
It stems from the substantial domain gap between standard VFI benchmarks and dual-camera zoom scenarios that feature larger viewpoint displacements and severer color variations.
A more principled approach is to collect specialized DCSZ datasets to retrain or fine-tune interpolation networks.
Unfortunately, capturing real-world continuous ground-truth sequences is physically intractable, as hardware constraints preclude deploying a dense array of sensors between the dual cameras.

To address this issue, we propose to assemble continuous virtual cameras and render intermediate views from a 3D scene to get DCSZ data, rather than relying on direct physical capture.
The formulation of such a virtual (V) camera requires accurate modeling of both geometric parameters (extrinsics and intrinsics) and photometric characteristics (\eg, the image signal processing (ISP) pipeline).
While extrinsic and intrinsic parameters define the perspective projection from 3D world coordinates to the 2D image plane, the ISP pipeline dictates the color rendering and tonal characteristics of the image.
Assuming that all parameters of both UW and W cameras are calibrated, virtual camera parameters can be obtained via smooth interpolation, allowing continuous views to be rendered from the underlying 3D scene.

In our preliminary conference work, ZoomGS~\cite{wu2024dual}, an initial step was taken toward virtual-camera-based DCSZ data synthesis. 
ZoomGS models the scene with 3D Gaussian Splatting (3DGS)~\cite{3DGS} and uses a multi-layer perceptron (MLP)-based camera transition (CamTrans) module to learn spatial and color transitions between the two endpoint 3DGS.
Nevertheless, ZoomGS suffers from two major limitations.
First, ZoomGS lacks intermediate-view supervision. This results in insufficient constraints on the CamTrans module that couples color and position, leading to temporal instability and visual jitter in the synthesized sequences.
Second, ZoomGS mandates dual-camera captures for the same scene, creating a data acquisition bottleneck that limits the scalability of dataset construction.

To overcome these challenges, we propose \textit{ZoomGS v2}, a novel and scalable data synthesis pipeline designed for high-fidelity DCSZ sequence generation.
Unlike ZoomGS, ZoomGS v2 explicitly decouples geometric and color transformations from 3D scene representation modeling.
Specifically, the framework separately estimates the relative camera pose and photometric discrepancy between the dual cameras, smoothly interpolates these attributes, and applies them during 3D scene rendering.
Crucially, benefiting from decoupling design, the underlying 3D scene representation to be reconstructed from standard single-camera multi-view data, which substantially improves data scalability.
Furthermore, previous VFI architectures struggle with the asymmetric correspondences and large geometric displacements between dual-camera views. To address this issue, we suggest ZoomFI, a VFI model specifically designed for the DCSZ task.
It leverages complementary bidirectional warping for comprehensive dual-camera information fusion.

To validate our approach, we collect real-world dual-camera image pairs from two mobile phones (Xiaomi and Huawei) for evaluation. 
Extensive experiments across multiple VFI backbones demonstrate that models fine-tuned on data synthesized by ZoomGS v2 achieve substantial quantitative and qualitative improvements over baseline models as well as ZoomGS, validating the effectiveness of the proposed introduced data synthesis pipeline. Moreover, the proposed ZoomFI consistently outperforms state-of-the-art VFI models on the evaluated DCSZ benchmarks.

The contributions can be summarized as follows:
\begin{itemize}
    \item We explore a practical task, \ie, dual-camera smooth zoom (DCSZ), and propose a solution that synthesizes continuous zoom sequences to empower video frame interpolation (VFI) models for smooth zoom previews.
    \item We propose \textit{ZoomGS v2}, a high-fidelity and scalable data synthesis framework that explicitly decouples geometric and color transformations from 3D scene representation modeling.
	\item We collect real-world dual-camera image pairs of two smartphones for evaluation. Extensive experiments validate the superiority of the proposed ZoomGS v2 and VFI model over existing state-of-the-art ones.
\end{itemize}

\section{Related Work}
\subsection{Frame Interpolation}
Given an image pair, video frame interpolation (VFI) aims to generate intermediate frames, which can be roughly divided into kernel-based methods~\cite{zhou2023exploring,kalluri2023flavr,shen2024ladder},
hallucination-based methods~\cite{kalluri2023flavr,kim2020fisr,shi2022video}, flow-based methods~\cite{RIFE,IFRNet,niklaus2020softmax,UPRNet,AMT,EMAVFI,VFIFormer,shen2024ladder,wu2024perception,guo2024generalizable,zhong2024clearer}, and state space model (SSM)-based ~\cite{zhang2024vfimamba}.
For example, RIFE~\cite{RIFE} and IFRNet~\cite{IFRNet} incorporate task-oriented flow distillation objectives to generate intermediate flow prior during training.
AMT~\cite{AMT} builds bidirectional correlation volumes for all pairs of pixels to retrieve correlations for more accurate flow estimation.
VFIMamba~\cite{zhang2024vfimamba} uses SSMs for multi-directional serialization of inter-frame spatiotemporal modeling and inter-frame features for generating intermediate frames.
In addition, some works~\cite{VFIFormer} introduce vision transformer~\cite{dosovitskiy2020image} for performance improvement.
However, the domain gap between DCSZ data and their training datasets limits their effective application for the DCSZ task. 
Moreover, compared to traditional interpolation, DCSZ interpolation has a larger amplitude of inter-frame motion, making it difficult for these traditional interpolation models to accurately model inter-frame motion.

\subsection{3D Reconstruction}
With the development of deep learning, neural rendering algorithms~\cite{NeRF,barron2021mip,yang2023freenerf,pumarola2021d,muller2022instant} have outperformed traditional methods~\cite{szeliski1996lumigraph,hanrahan1996light} in 3D reconstruction.
Neural Radiance Field (NeRF)~\cite{NeRF} pioneers this field and facilitates high-quality novel view synthesis with the use of MLPs.
Advancements have been made to handle multi-resolution inputs~\cite{barron2021mip}, sparse-view image inputs~\cite{yang2023freenerf}, dynamic scenes~\cite{pumarola2021d}, and nearly real-time rendering capabilities~\cite{muller2022instant}.
In addition, a few works~\cite{tosi2023nerf,tosi2023nerf,yen2022nerf} exploit the 3D priors in NeRF models to construct data for downstream tasks, such as stereo depth estimation~\cite{tosi2023nerf}, object detection~\cite{ge2022neural}, and object dense description~\cite{yen2022nerf}.
However, the NeRF-based 3D reconstruction methods struggle to achieve visually realistic rendering quality and real-time rendering speed simultaneously, due to the implicit scene representation and the dense sampling of rays.

In contrast, the recent 3D Gaussian Splatting (3DGS)~\cite{3DGS} overcomes the issue with explicit scene representation and differentiable rasterization, attracting substantial attention~\cite{chen2024survey}.
For example, some works utilize it for 3D reconstruction in sparse-view scenarios~\cite{chung2023depth,xiong2023sparsegs,yang2024gaussianobject,zhu2023fsgs,szymanowicz2023splatter,charatan2023pixelsplat,zhang2024cor,chan2024point,wu2024dig3d,zhu2024fsgs}, dynamic scenarios~\cite{wu20244d,li2024spacetime,wu20244d,liu2024dynamic,wu2026deblur4dgs}, and COLMAP-free scenarios~\cite{fu2023colmap,chen2024zerogs,wei2025pcr,fu2024colmap}.
Some works employ it in segmentation~\cite{hu2024semantic,cen2023segment,lan20232d,jain2024gaussiancut}, visual generation~\cite{melas20243d,pan2024fast,ren2023dreamgaussian4d,chen2024text,li2025controllable} and editing~\cite{zhuang2024tip,fang2023gaussianeditor,feng2024evsplitting,fan2024trim,wan2026color3d} tasks. 
Besides, many efforts have been made in anti-aliased rendering~\cite{yu2023mip,yan2023multi} and storage efficiency~\cite{fan2023lightgaussian,papantonakis2024reducing,zhang2024gaussianforest}.
In this work, further considering the specific imaging characteristics for each virtual camera, we enhance 3DGS~\cite{3DGS} with the ability to specialize the 3D model for a specific camera, thus generating data for the DCSZ task.

\section{Proposed Problem and Method}
In this section, we first formulate the problem in Sec.~\ref{section:4_0_2} and present the motivation of our ZoomGS v2 method in Sec.~\ref{section:4_0_3}.
We then introduce the proposed ZoomGS v2 in Sec.~\ref{section:4_2_2}, Sec.~\ref{section:4_2_3} and Sec.~\ref{section:4_2_4}.
Finally, we detail the proposed ZoomFI and provide the training details in Sec.~\ref{section:4_4}.

\subsection{Problem DCSZ Setup} 
\label{section:4_0_2}
% \ 
% \newline
%
As illustrated in Fig.~\ref{fig:intro}(c), the distinct deployment locations and internal characteristics~\cite{zhang2022self,alzayer2023dc2,abdelhamed2021leveraging} of dual-zoom cameras lead to prominent preview discontinuities in geometric perspective and color rendition, which significantly affect the user’s zoom experience.
The proposed dual-camera smooth zoom (DCSZ) task aims to achieve a fluid zoom preview, which can be accomplished by employing a video frame interpolation (VFI) model to synthesize $N$ intermediate images $\{\tilde{\mathbf{{I}}}_{i}\}_{i=1}^{N}$ between dual-camera images. 
Denoted by $\mathbf{I}_{uw}$ and $\mathbf{I}_{w}$ the UW and W images respectively, the VFI model can be written as:
\begin{align}
	\tilde{\mathbf{{I}}}_{i} = \mathcal{I}(\mathbf{I}_{uw}, \mathbf{I}_{w}, t_{i};\mathrm{\Theta_{\mathcal{F}}}),
\end{align}
where $\mathrm{\Theta_{\mathcal{F}}}$ denotes the parameters of the VFI model $\mathcal{F}$. 
$t_{i}$ is the temporal encoding that indicates the relative relationship between the synthesized image and the dual-camera images.
Due to the substantial domain gap between standard VFI benchmarks and dual-camera zoom scenarios, existing pre-trained VFI models often fail to meet expectations.
Retraining or fine-tuning the model is a more appropriate solution, but it is impractical to capture the continuous ground-truth sequences by deploying a dense array of sensors between the dual cameras.

To address the issue, we suggest assembling continuous virtual cameras to construct such zoom sequences.
For each dual-camera image pair, we construct a set of $N$ virtual cameras $\{v_{i}\}^{N}_{i=1}$ along the transition between the UW and W cameras.
Thus, synthetic ground-truth $\{{\mathbf{{I}}}_i\}^{N}_{i=1}$ can be rendered by these virtual cameras.
By performing the steps on multiple image pairs, a synthetic dataset for the DCSZ task can be constructed, thus being employed to fine-tune a VFI model, \ie,
\begin{align}
	\label{eq:VFI_training_eqn}
	\mathrm{\Theta_{\mathcal{I}}^*} = 
	\arg \min _{\mathrm{\Theta_{\mathcal{F}}}} \mathcal{L}_{VFI}
	\left(\mathcal{I}
	(\mathbf{I}_{uw}, \mathbf{I}_{w},
	t_{i}; 
	\mathrm{\Theta_{\mathcal{F}})}, 
	{\mathbf{{I}}}_i\right),
\end{align}
where $\mathcal{L}_{VFI}$ denotes the learning objective during training.

To realize this, we construct virtual cameras comprising two key components. 
The first component consists of the camera extrinsic and intrinsic parameters, as they determine the perspective projection from 3D world coordinates to the 2D image plane.
The second component comprises the ISP pipeline parameters, which influence the color rendering and tonal characteristics of the image.
Assuming that all parameters of UW and W cameras have been calibrated, virtual camera parameters can be obtained through smooth interpolation to render continuous views from the underlying 3D scene.

\begin{figure*}[t!]
	\centering
	\includegraphics[width=1.0\linewidth]{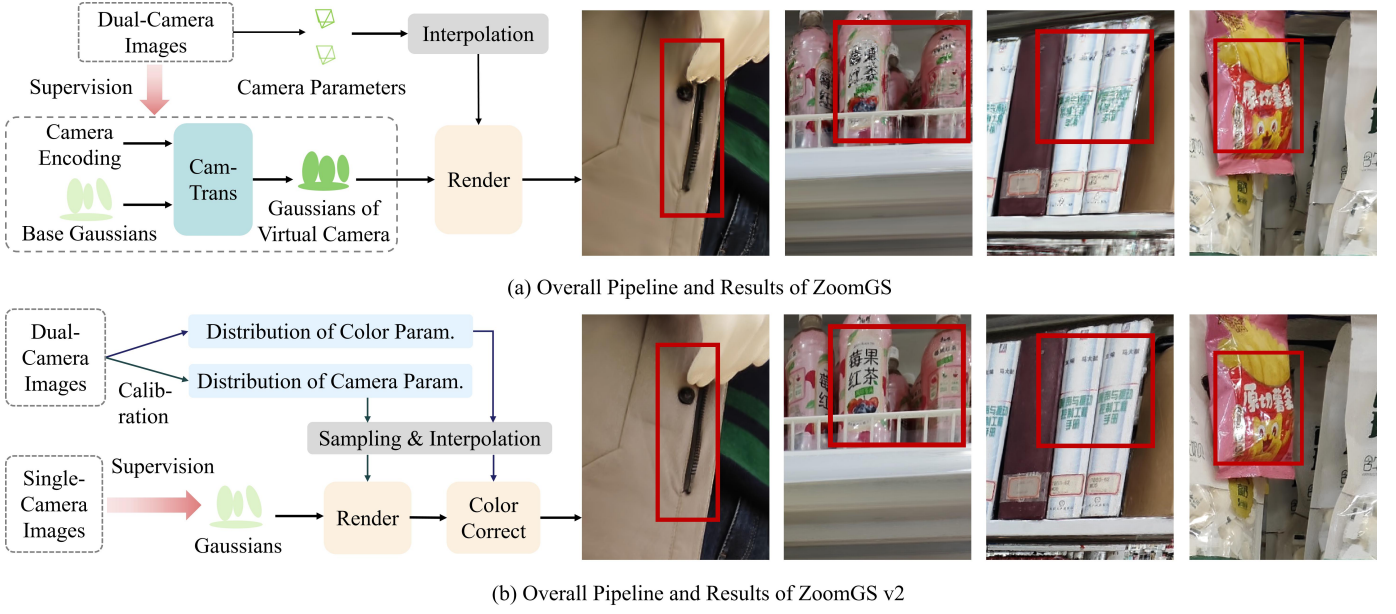}
	% \vspace{-3mm}
	\caption{
		Comparison of ZoomGS~\cite{wu2024dual} and ZoomGS v2.
        (a) Overall Pipeline and Results of ZoomGS. ZoomGS models the scene with Base Gaussians and uses the CamTrans Module to learn spatial and color transitions between the two endpoint 3DGS~\cite{3DGS} representations, and then transforms the Base Gaussians to target Gaussians of virtual cameras for rendering.
        (b) Overall Pipeline and Results of ZoomGS v2. ZoomGS v2 first calibrates the distribution of camera and color parameters from a small set of dual-camera data, and then applies them to Gaussians constructed from large-scale single-camera data to generate 3DGS representations for scene rendering.
		}
	\label{Fig:pipline_compare}
	% \vspace{-3mm}
    \vspace{0mm}
\end{figure*}

\subsection{Motivation of Methods} 
\label{section:4_0_3}
% \ 
Our preliminary conference work, ZoomGS~\cite{wu2024dual}, attempted to address these issues by constructing 3D models using 3D Gaussian Splatting (3DGS)~\cite{3DGS}. ZoomGS represents the 3D scenes of the UW and W cameras with 3DGS, and then utilizes an MLP-based camera transition (CamTrans) module to learn the position and color transformations of 3DGS representations between two endpoints, ultimately interpolating them through CamTrans to obtain virtual camera representations, as illustrated in Fig.~\ref{Fig:pipline_compare} (a).
However, ZoomGS lacks intermediate-view supervision. This results in insufficient constraints on the CamTrans module that couples color and position together, leading to discontinuous intermediate sequences (see Fig.~\ref{Fig:pipline_compare}(a)). Moreover, ZoomGS requires dual-camera data from the same scene, rendering scalable dataset expansion prohibitively costly, as existing open-source 3D Gaussian training datasets are captured by a single camera.

\begin{figure}[t!] % [t]
	\centering
	\includegraphics[width=0.99\linewidth]{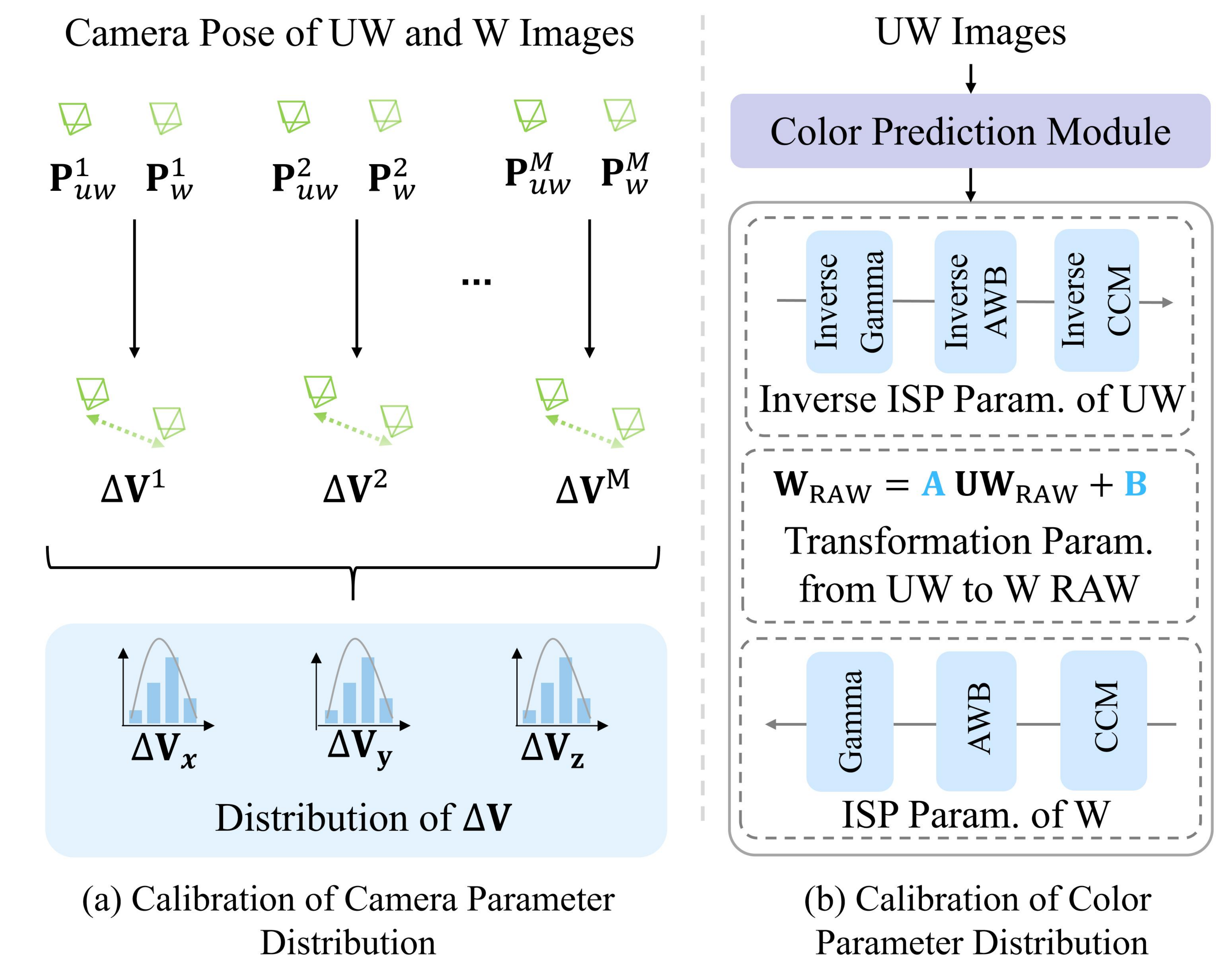}
	\vspace{-2mm}
    \caption{
        (a) Calibration of Camera Parameter Distribution. ZoomGS v2 collects multiple camera parameter offsets $\Delta \mathbf{V}$ (Eq.~\eqref{equ32303}) from UW-W pairs and models their distribution using a Gaussian distribution.
		(b) Calibration of Color Parameter Distribution. ZoomGS v2 trains a CNN-based~\cite{kim2014convolutional} color prediction module on UW-W pairs to predict the parameters of the color transformation pipeline from UW to W.
	}
	\label{fig:pipeline_2}
	\vspace{0mm}
\end{figure}

These observations motivate us to disentangle scene representation learning from device-specific camera transformation modeling.
To this end, we propose a novel and scalable synthesis pipeline termed ZoomGS v2 (see Fig.~\ref{Fig:pipline_compare}(b)). 
Specifically, ZoomGS v2 first calibrates the geometric and color transformation gap between dual cameras from a small amount of dual-camera calibration data.
Geometric transformation is modeled by Gaussian distribution statistics of dual-camera parameter gap, and color transformation is modeled by image signal processing (ISP) pipeline with a network introduced and trained to predict these parameters, as shown in Fig.~\ref{fig:pipeline_2}.
Then, ZoomGS v2 transfers the differences to large-scale single-camera scenes, and simultaneously models 3DGS scene with single-camera data,
instead of jointly optimizing scene representation and cross-camera transformation from sparse dual-camera observations.
%
% Specifically, ZoomGS v2 first randomly selects an image with its camera parameters from the single camera data to serve as the UW view. Then it calculates corresponding W camera parameters through the geometric transformation process, as well as predicts the parameters of the color transformation with the CNN. Finally ZoomGS v2 calculates the camera parameters and color parameters of virtual cameras by interpolating those of UW and W, enabling the DCSZ data to be rendered from single camera 3D model.
% %
%
By decoupling scene representation learning from cross-camera transformation modeling, ZoomGS v2 avoids the lack of intermediate-view supervision in ZoomGS and enables scalable synthesis of DCSZ training data from large-scale single-camera datasets.

We divide the introduction of ZoomGS v2 into three parts: Geometric Transformation Calibration of Dual-camera in Sec.~\ref{section:4_2_2}, Color Transformation Prediction of Dual-camera in Sec.~\ref{section:4_2_3}, and Data Generation from ZoomGS v2 in Sec.~\ref{section:4_2_4}. 
In addition, to address the asymmetric correspondences and large geometric displacements between dual-camera views faced by VFI models in the DCSZ task, we design an effective VFI model tailored for the DCSZ task, termed ZoomFI, which incorporates both forward and backward warping to produce more photorealistic zoom transitions, as detailed in Sec.~\ref{section:4_4}.

\subsection{Geometric Transformation Calibration of Dual-camera} 
\label{section:4_2_2}
% \
% \newline
The geometric disparity between the UW and W images can be primarily reflected by their camera parameters, including both intrinsic and extrinsic parameters. 
Since 3DGS~\cite{3DGS} enables rendering directly from camera parameters, this disparity can be characterized by analyzing the variations in these parameters.
The camera extrinsic parameters $\mathbf{P} = [\mathbf{R}\mid\mathbf{T}]$ consist of a rotation matrix $\mathbf{R} \in \mathbb{R}^{3\times 3}$ and a translation vector $\mathbf{T} \in \mathbb{R}^{3\times 1}$, which encode the camera's orientation and position in world coordinates, respectively.
The intrinsic parameters include the focal length, which determines the field of view (FOV) of the captured images.
For clarity, in the following sections, we denote a pair of dual-camera images as $\mathbf{I}_{uw}$ and $\mathbf{I}_{w}$, the corresponding extrinsic parameters as $\mathbf{P}_{uw} = [\mathbf{R}_{uw}|\mathbf{T}_{uw}]$ and $\mathbf{P}_{w} = [\mathbf{R}_{w}|\mathbf{T}_{w}]$, and their intrinsic parameters as $\mathbf{K}_{uw}$ and $\mathbf{K}_{w}$, respectively.

Camera intrinsics are primarily determined by the focal length. 
For a smartphone, the effective focal length ratio (or scaling factor) between the UW and W cameras is a fixed constant $s$, yielding a constant relative intrinsic disparity. 
This relationship can be formulated as, $\mathbf{K}_{w} = \mathbf{K}_{uw} / s$.

\begin{figure}[t!]
	\centering
    \hfill
	\includegraphics[width=3.3in]{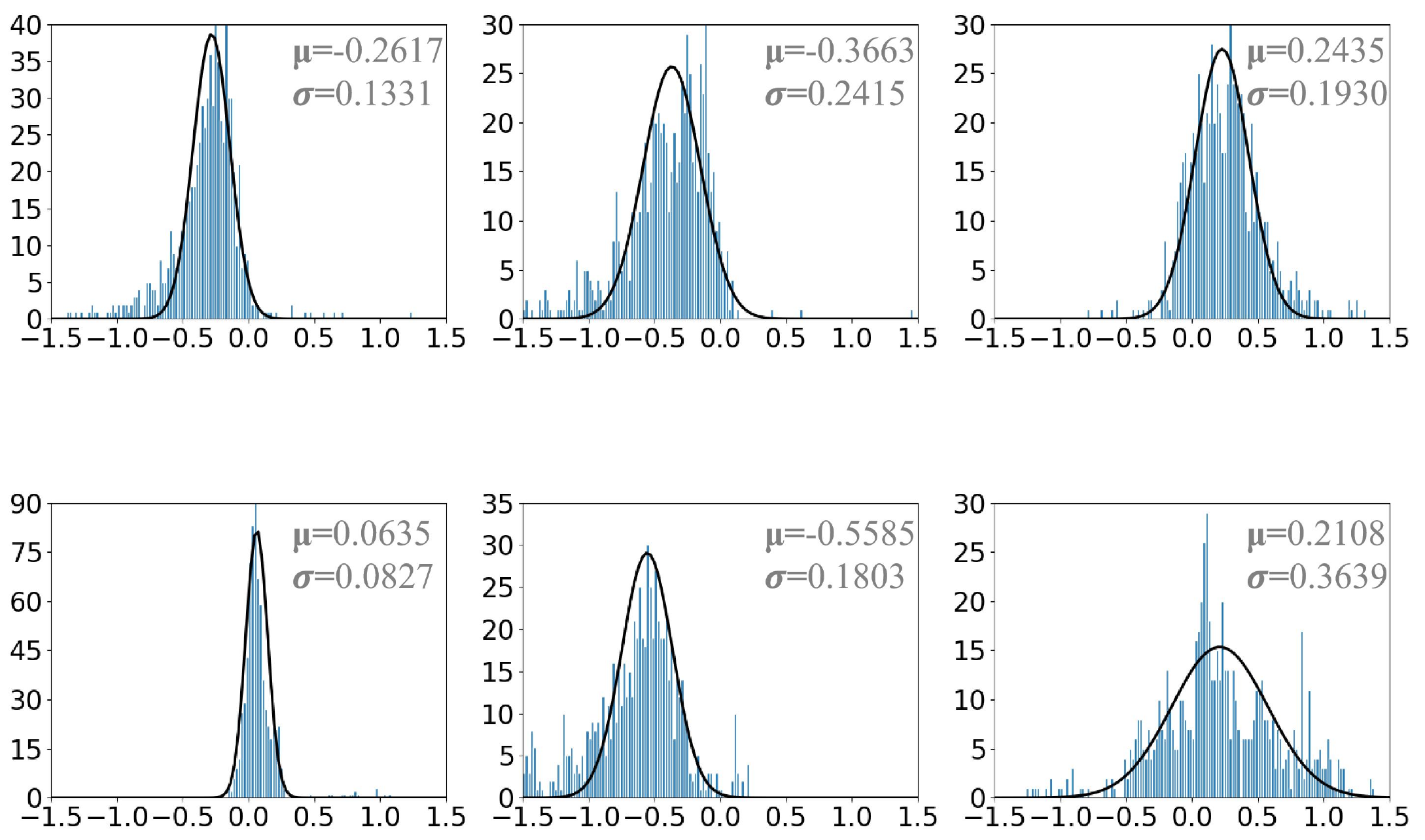}%
    \put(-246,86){{\fontsize{7.5}{7}\selectfont \rotatebox{90}{Num. of samples}}}
    \put(-208,75){{\fontsize{7.5}{7}\selectfont $\Delta \mathbf{V}_{x}$}}
    \put(-128,75){{\fontsize{7.5}{7}\selectfont $\Delta \mathbf{V}_{y}$}}
    \put(-49,75){{\fontsize{7.5}{7}\selectfont $\Delta \mathbf{V}_{z}$}}
    \put(-188,65){{\fontsize{7.5}{7}\selectfont (a) Distribution of $\Delta \mathbf{V}$ for Xiaomi smartphone}}
    \put(-246,6){{\fontsize{7.5}{7}\selectfont \rotatebox{90}{Num. of samples}}}
    \put(-208,-5){{\fontsize{7.5}{7}\selectfont $\Delta \mathbf{V}_{x}$}}
    \put(-128,-5){{\fontsize{7.5}{7}\selectfont $\Delta \mathbf{V}_{y}$}}
    \put(-49,-5){{\fontsize{7.5}{7}\selectfont $\Delta \mathbf{V}_{z}$}}
    \put(-188,-15){{\fontsize{7.5}{7}\selectfont (b) Distribution of $\Delta \mathbf{V}$ for Huawei smartphone}}
	\vspace{2mm}
    \caption{Statistical distributions and Gaussian fitting results of $\Delta \mathbf{V}$, which represents the relative vector offset between the UW and W images captured by Xiaomi (a) and Huawei (b) model smartphones. From left to right, the plots represent $\Delta \mathbf{V}_{x}$, $\Delta \mathbf{V}_{y}$ and $\Delta \mathbf{V}_{z}$, respectively.}
	\label{fig_trans}
    \vspace{0mm}
\end{figure}
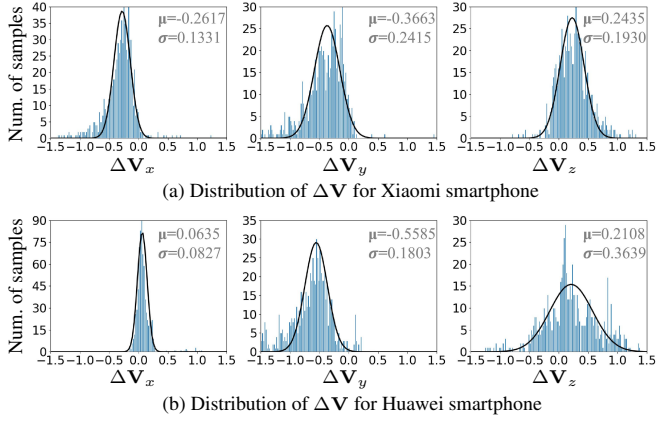

To formulate the spatial relationship between the translation vectors $\mathbf{T}_{uw}$ and $\mathbf{T}_{w}$ of the dual cameras, we introduce a relative spatial displacement vector $\Delta \mathbf{V}$.
Without loss of generality, the upright orientation of the mobile device is established as the reference coordinate frame.
Owing to the rigid physical mounting of the dual-camera module on the mobile chassis, the relative spatial displacement between the two optical centers remains invariant within the coordinate frame of the device.
Accordingly, the relative displacement vector $\Delta \mathbf{V}$ defined in the world coordinate system is formulated as:
\begin{align}
	\label{equ32303}
	\Delta \mathbf{V} = \mathbf{R}^{\mathsf{T}}_{w} \mathbf{T}_{w} - \mathbf{R}^{\mathsf{T}}_{uw} \mathbf{T}_{uw}.
\end{align}

Regarding the rotation matrices $\mathbf{R}_{uw}$ and $\mathbf{R}_{w}$, because the dual sensors are rigidly fixed to the same planar substrate, their relative orientation discrepancy is marginal in practical scenarios.
Empirical statistical analysis on our collected dual-camera calibration dataset reveals that the mean absolute error (MAE) between the rotation parameters of the UW and W cameras is less than $3.5 \times 10^{-3}$.
Consequently, we reasonably approximate the orientations of both cameras as identical during data synthesis, \ie, $\mathbf{R}_{w} \approx \mathbf{R}_{uw}$.

To determine the empirical distribution of the relative displacement vectors $\Delta \mathbf{V}$, we utilize a diverse set of calibrated dual-camera image pairs along with their estimated camera extrinsics.
Theoretically, the relative vector offset $\Delta \mathbf{V}$ should remain deterministic for a specific mobile device model.
In practice, however, perturbations induced by subtle camera shake, sensor thermal variations, and numerical ambiguities inherent to Structure-from-Motion (SfM) estimation in COLMAP~\cite{schonberger2016structure} introduce stochastic fluctuations into the calculated values of $\Delta \mathbf{V}$.
To account for this stochasticity, we model $\Delta \mathbf{V}$ as a multivariate Gaussian distribution fitted over the empirical measurements, where each coordinate component of the translation vector comprises a deterministic baseline offset perturbed by Gaussian noise (see Fig.~\ref{fig_trans}).

Thus, given an arbitrary single-camera UW view along with its extrinsic parameters $\mathbf{R}_{uw}$ and $\mathbf{T}_{uw}$, a relative displacement vector $\Delta \mathbf{V}$ can be randomly sampled from the fitted multivariate Gaussian distribution.
Under the approximation $\mathbf{R}_{w} \approx \mathbf{R}_{uw}$, the corresponding translation vector $\mathbf{T}_{w}$ of the W camera is analytically derived via Eq.~\eqref{equ32303} as $\mathbf{T}_{w} \approx \mathbf{R}_{w} (\Delta \mathbf{V} + \mathbf{R}^{\mathsf{T}}_{uw} \mathbf{T}_{uw})$. In summary, the extrinsic parameters of the W camera can be systematically derived from those of the UW camera through the proposed decoupled geometric formulation.

% \vspace{-3mm}
\subsection{Color Transformation Prediction of Dual-camera} 
\label{section:4_2_3}
% \
% \newline

To model the photometric disparity between the dual cameras, we leverage the representative image signal processing (ISP) pipeline of mobile devices to formulate the color transformation between the RGB images $\mathbf{I}_{uw}, \mathbf{I}_{w}$ and their underlying linear RAW representations $\mathbf{X}_{uw}, \mathbf{X}_{w}$, as illustrated in Fig.~\ref{fig:pipeline_2}(b).
Conceptually, mapping the color distribution of $\mathbf{I}_{uw}$ to that of $\mathbf{I}_{w}$ involves a tripartite workflow: first inverting $\mathbf{I}_{uw}$ into the linear RAW domain $\mathbf{X}_{uw}$ via an inverse ISP, subsequently mapping $\mathbf{X}_{uw}$ to $\mathbf{X}_{w}$ through a linear transformation, and finally rendering $\mathbf{X}_{w}$ into the RGB domain $\mathbf{I}_{w}$ via a forward ISP.

Specifically, a linear transformation parameterized by a weight matrix $\mathbf{W}_{raw} \in \mathbb{R}^{3 \times 3}$ and a bias vector $\mathbf{b}_{raw} \in \mathbb{R}^{3 \times 1}$ is initially employed to characterize the cross-sensor spectral discrepancy from $\mathbf{X}_{uw}$ to $\mathbf{X}_{w}$.
We then simplify the ISP pipeline into three primary stages governing color rendition, including color correction (CC), auto white balance (AWB), and gamma correction.
We define full matrices $\mathbf{C}_{uw}, \mathbf{C}_{w} \in \mathbb{R}^{3 \times 3}$ to model the CC transformations, and diagonal matrices $\mathbf{D}_{uw}, \mathbf{D}_{w} \in \mathbb{R}^{3 \times 3}$ to characterize their respective AWB gains of the UW and W cameras, respectively.
The non-linear gamma corrections in the two ISP pipelines are formulated channel-wise as:
\begin{align}
    \label{equgammauw}
    \mathbf{I}_{uw,c} = \frac{(\alpha_{uw,c} \mathbf{I}_{uw,c}^{awb} + \beta_{uw,c})^{\gamma_{uw,c}} - (\beta_{uw,c})^{\gamma_{uw,c}}}{(\alpha_{uw,c} + \beta_{uw,c})^{\gamma_{uw,c}} - (\beta_{uw,c})^{\gamma_{uw,c}} },
\end{align}
\begin{align}
    \label{equgammaw}
    \mathbf{I}_{w,c} = \frac{(\alpha_{w,c} \mathbf{I}_{w,c}^{awb} + \beta_{w,c})^{\gamma_{w,c}} - (\beta_{w,c})^{\gamma_{w,c}}}{(\alpha_{w,c} + \beta_{w,c})^{\gamma_{w,c}} - (\beta_{w,c})^{\gamma_{w,c}} },
\end{align}
where $c \in \{r, g, b\}$ denotes the color channel, $\{\alpha_{uw,c}, \beta_{uw,c}, \gamma_{uw,c}\}$ and $\{\alpha_{w,c}, \beta_{w,c}, \gamma_{w,c}\}$ are parameters to be determined, and $\mathbf{I}_{uw}^{awb}, \mathbf{I}_{w}^{awb}$ represent the intermediate representations obtained after applying CC and AWB to $\mathbf{X}_{uw}$ and $\mathbf{X}_{w}$, respectively.
Consequently, the intermediate linear operations can be algebraically unified. By defining $\mathbf{W} = \mathbf{D}_{w}\mathbf{C}_{w}\mathbf{W}_{raw}\mathbf{C}_{uw}^{-1}\mathbf{D}_{uw}^{-1} \in \mathbb{R}^{3 \times 3}$ and $\mathbf{b} = \mathbf{D}_{w}\mathbf{C}_{w}\mathbf{b}_{raw} \in \mathbb{R}^{3 \times 1}$, the color transition from $\mathbf{I}_{uw}^{awb}$ to $\mathbf{I}_{w}^{awb}$ is jointly expressed as a compact affine mapping, \ie, $\mathbf{I}_{w}^{awb} = \mathbf{W}\mathbf{I}_{uw}^{awb} + \mathbf{b}$.

Because these physical parameters are inaccessible from closed-source mobile imaging systems, we design a learning-based estimation strategy.
We collect a diverse set of real-world UW and W image pairs and establish geometric alignment for each pair via field-of-view cropping followed by optical-flow-based warping~\cite{9722682}.
A convolutional neural network (CNN)~\cite{kim2014convolutional} is then trained on these aligned pairs to directly predict the transformation parameters.
The optimization is supervised by a color difference loss based on $\mathrm{\Delta E}$~\cite{mclaren1976xiii}, which projects the predicted and ground-truth images into the perceptually uniform CIELAB color space and computes their $\mathrm{L}_{2}$ Euclidean distance.
During inference, the estimated photometric parameters are applied to transform the color profile of the UW view into that of the W camera.

\begin{figure}[t!]
	\centering
    \begin{overpic}[width=0.48\textwidth,grid=False]
        {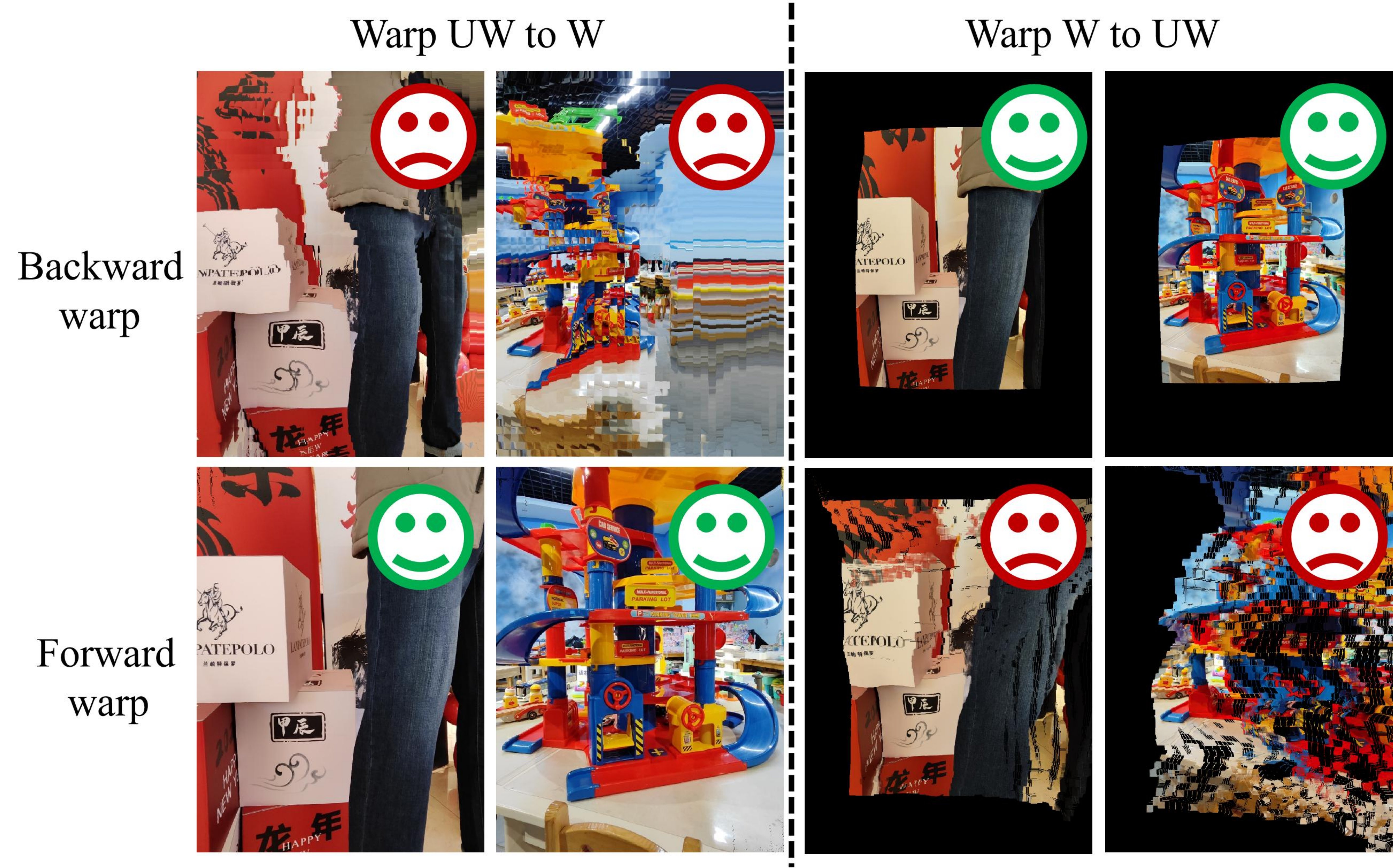}
    \end{overpic}
    \vspace{2mm}
	\caption{Comparison of forward and backward warping for transforming UW to W (Left) and transforming W to UW (Right) using SEA-RAFT~\cite{searaft}.
 	} 
	\label{fig_fprwardflow}
	\vspace{0mm}
\end{figure}

\subsection{Data Generation from ZoomGS v2} 
\label{section:4_2_4}
% \
% \newline

To synthesize continuous dual-camera zoom sequences, we curate diverse static scenes from DL3DV-10K~\cite{dl3dv} and reconstruct each scene as a 3DGS representation.
For each reconstructed scene, ZoomGS v2 first randomly samples an original capture to define the reference UW view, with corresponding intrinsic matrix $\mathbf{K}_{uw}$ and extrinsic matrix $\mathbf{P}_{uw}=[\mathbf{R}_{uw} \mid \mathbf{T}_{uw}]$.
The target W camera intrinsics $\mathbf{K}_{w}$ and orientation $\mathbf{R}_{w}$ are obtained according to Sec.~\ref{section:4_2_2}, while the translation vector $\mathbf{T}_{w}$ is determined by:
\begin{align}
	\label{equ465r8457}
	\mathbf{T}_{w} = \mathbf{T}_{uw} + \mathbf{R}_{w} \Delta \mathbf{V},
\end{align}
where $\Delta \mathbf{V} \in \mathbb{R}^{3 \times 1}$ is drawn from the fitted multivariate Gaussian distribution.

To simulate smooth camera movement between the two endpoints, we define a normalized progression factor $\tau_i = \frac{i-1}{N-1} \in [0, 1]$ for the $i$-th virtual camera view ($i \in \{1, \dots, N\}$).
Following the continuous trajectory modeling in~\cite{liu2021mba,wang2023bad}, the intermediate extrinsic parameters $\mathbf{P}_{v_{i}} \in \mathrm{SE}(3)$ are computed via geodesic interpolation in the Lie algebra of $\mathfrak{se}(3)$:
\begin{align}
	\mathbf{P}_{v_{i}} = \mathbf{P}_{uw} \exp \left( \tau_i \log \left( \mathbf{P}_{w}\mathbf{P}_{uw}^{-1} \right) \right),
\end{align}
where $\exp(\cdot)$ and $\log(\cdot)$ denote the matrix exponential and logarithmic mappings between the Lie group $\mathrm{SE}(3)$ and its Lie algebra $\mathfrak{se}(3)$, respectively.
Simultaneously, the intrinsic parameters are smoothly interpolated as:
\begin{align}
	\mathbf{K}_{v_{i}} = (1 - \tau_i) \mathbf{K}_{uw} + \tau_i \mathbf{K}_{w}.
\end{align}

With the interpolated camera parameters $(\mathbf{K}_{v_i}, \mathbf{P}_{v_i})$, the intermediate geometric view $\mathbf{I}_{v_{i}}$ is rendered from the reconstructed 3DGS representation.
Finally, the photometric transformation parameters predicted by the CNN~\cite{kim2014convolutional} model are linearly modulated by $\tau_i$ and applied to $\mathbf{I}_{v_{i}}$ via the formulation in Sec.~\ref{section:4_2_3}, thereby generating photometrically and geometrically continuous zoom sequences.

\begin{figure}[t!]
	\centering
	\includegraphics[width=3.2in]{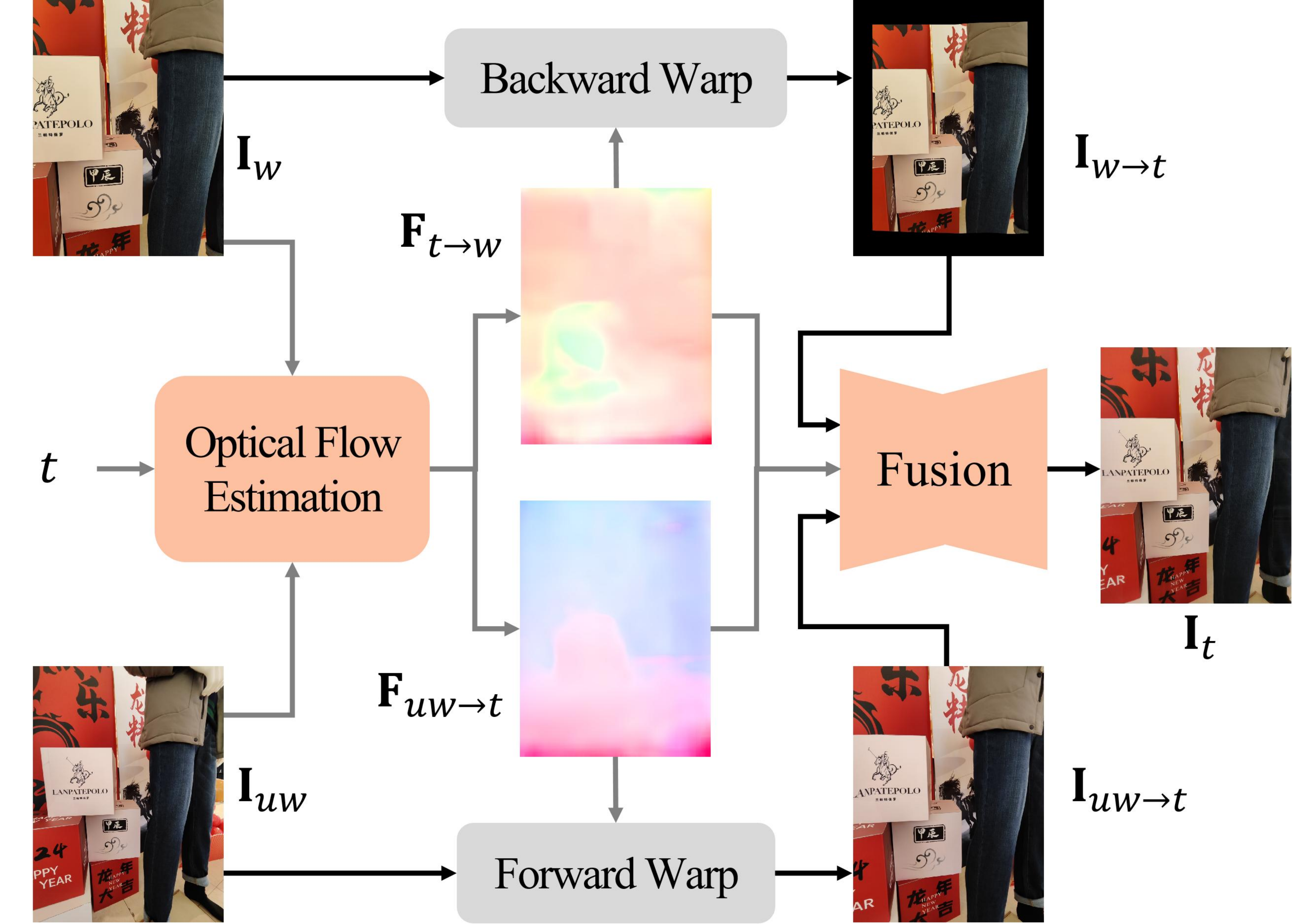}%
	\vspace{2mm}
    \caption{Overview pipeline of ZoomFI. The bidirectional warping is used for dual-camera information fusion.}
	\label{fig_model}
    \vspace{0mm}
\end{figure}

\subsection{Pipeline and Training of ZoomFI}
\label{section:4_4}

The DCSZ task departs substantially from conventional video frame interpolation (VFI) in that it entails significantly larger motion amplitudes, while the field of view (FoV) of the UW camera strictly subsumes that of the W camera.
Conventional flow-based VFI architectures (\eg, RIFE~\cite{RIFE} and AMT~\cite{AMT}) typically estimate backward optical flows from the intermediate target frame to the input frames, backward warp the input images toward the target time step, and fuse the aligned representations to generate the intermediate view.
However, because backward warping executes target-oriented pixel sampling, the substantial scale disparity and large spatial displacement between dual-camera views often induce sampling ambiguities and severe peripheral distortions when backward warping the broad-FoV UW view toward the localized W view (see Fig.~\ref{fig_fprwardflow}).
To address this limitation, we avoid relying exclusively on backward warping and instead synergistically couple forward warping with backward warping.
Because forward warping operates along the inverse sampling trajectory, it inherently accommodates scale expansion and preserves peripheral visual content, thereby providing information complementary to backward warping for DCSZ tasks.
Furthermore, a specialized optical flow estimation network is integrated to robustly handle the substantial scale discrepancy between the dual-camera views.

Specifically, we utilize a pre-trained optical flow network $\mathcal{O}$ (\eg, SEA-RAFT~\cite{searaft}) to compute bidirectional flow fields between the input images $\mathbf{I}_{uw}$ and $\mathbf{I}_{w}$ at a $4\times$ downsampled resolution:
\begin{align}
	\label{equ311}
	\mathbf{F}_{uw\rightarrow w} = \mathcal{O}(\mathbf{I}_{uw}, \mathbf{I}_{w}), \quad \mathbf{F}_{w\rightarrow uw} = \mathcal{O}(\mathbf{I}_{w}, \mathbf{I}_{uw}).
\end{align}
Following the flow approximation scheme in SGM-VFI ~\cite{liu2024sparse}, intermediate backward flows $\mathbf{F}_{i\rightarrow uw}$ and $\mathbf{F}_{i\rightarrow w}$ are approximated via forward warping, while intermediate forward flows $\mathbf{F}_{uw\rightarrow i}$ and $\mathbf{F}_{w\rightarrow i}$ are derived through linear trajectory interpolation, providing a coarse motion initialization:
\begin{subequations}\label{equ31345}
\begin{align}
	\mathbf{F}_{i\rightarrow uw} &= \overrightarrow{\mathcal{W}}\big(\tau_i \mathbf{F}_{w\rightarrow uw}, (1-\tau_i) \mathbf{F}_{w\rightarrow uw}\big),\label{equ313}\\
    \mathbf{F}_{i\rightarrow w} &= \overrightarrow{\mathcal{W}}\big((1-\tau_i) \mathbf{F}_{uw\rightarrow w}, \tau_i \mathbf{F}_{uw\rightarrow w}\big),\label{equ314}\\
    \mathbf{F}_{uw\rightarrow i} &= \tau_i \mathbf{F}_{uw\rightarrow w}, \quad \mathbf{F}_{w\rightarrow i} = (1-\tau_i) \mathbf{F}_{w\rightarrow uw},\label{equ315}
\end{align}
\end{subequations}
where $\overrightarrow{\mathcal{W}}(\mathbf{X}, \mathbf{F})$ denotes the forward warping operator that splats tensor $\mathbf{X}$ guided by the flow field $\mathbf{F}$.

The overview pipeline of ZoomFI can be seen in Fig.~\ref{fig_model}. 
The backward flows $\mathbf{F}_{i\rightarrow uw}$ and $\mathbf{F}_{i\rightarrow w}$ are employed to backward warp $\mathbf{I}_{uw}$ and $\mathbf{I}_{w}$ to the intermediate state $i$, yielding warped frames $\mathbf{W}_{i\rightarrow uw}$ and $\mathbf{W}_{i\rightarrow w}$, respectively.
Simultaneously, the forward flows $\mathbf{F}_{uw\rightarrow i}$ and $\mathbf{F}_{w\rightarrow i}$ forward warp $\mathbf{I}_{uw}$ and $\mathbf{I}_{w}$ to state $i$, yielding $\mathbf{W}_{uw\rightarrow i}$ and $\mathbf{W}_{w\rightarrow i}$.
Subsequently, the aligned feature tensors $\{\mathbf{W}_{i\rightarrow uw}, \mathbf{W}_{i\rightarrow w}, \mathbf{W}_{uw\rightarrow i}, \mathbf{W}_{w\rightarrow i}\}$ and the intermediate flow fields $\{\mathbf{F}_{i\rightarrow uw}, \mathbf{F}_{i\rightarrow w}, \mathbf{F}_{uw\rightarrow i}, \mathbf{F}_{w\rightarrow i}\}$ are concatenated along the channel dimension and fed into a multi-scale fusion module to synthesize the final intermediate frame $\tilde{\mathbf{I}}_{i}$.
The fusion module consists of initial convolutional feature extraction layers followed by a multi-scale U-Net architecture.

Furthermore, to facilitate real-time deployment on resource-constrained mobile hardware, we introduce an efficient variant termed ZoomFI-Lite.
By executing optical flow estimation at an $8\times$ downsampling factor, ZoomFI-Lite achieves accelerated inference throughput and a reduction in floating-point operations (FLOPs) compared with the standard ZoomFI architecture.

To optimize the proposed interpolation framework, we employ a joint loss combining a multi-scale Laplacian pyramid reconstruction loss and a perceptual loss.
Following established VFI methodologies~\cite{niklaus2020softmax,RIFE}, the reconstruction loss is computed over the Laplacian pyramid representations of the predicted frame $\tilde{\mathbf{I}}_{i}$ and the ground-truth frame $\mathbf{I}_{i}$.
The overall optimization objective $\mathcal{L}_{\mathrm{VFI}}$ is formulated as:
\begin{align}
	\label{eq:VFI_loss}
	\mathcal{L}_{\mathrm{VFI}} = \mathcal{L}_{1}\big(\mathrm{LP}(\tilde{\mathbf{I}}_{i}), \mathrm{LP}(\mathbf{I}_{i})\big) + \lambda_{\mathrm{per}} \mathcal{L}_{1}\big(\phi(\tilde{\mathbf{I}}_{i}), \phi(\mathbf{I}_{i})\big),
\end{align}
where $\lambda_{\mathrm{per}}$ is a hyperparameter set to 1.0, $\mathrm{LP}(\cdot)$ designates the Laplacian pyramid decomposition, and $\phi(\cdot)$ denotes the features extracted from a pre-trained VGG-19 network~\cite{simonyan2014very}.

\section{Experiments}

To evaluate the generalization capability of our method across different devices and zooming paths from UW (zoom factor $\times0.6$) to W (zoom factor $\times1.0$), we conduct experiments using two smartphones from Xiaomi and Huawei, respectively. And we consider two zooming settings: $\times0.6\leftrightarrow\times1.0$ and $\times0.85\leftrightarrow\times1.0$ 
In the $\times0.6\leftrightarrow\times1.0$ setting, we directly apply continuous geometric and color transformations from the UW view to the W view. In the $\times0.85\leftrightarrow\times1.0$ setting, we first apply center cropping to the UW image to obtain an intermediate view with a zoom factor close to that of the W view (\ie, $\times0.85$), and then apply continuous geometric and color transformations from this intermediate view to the W view.
We construct training and test datasets for both devices in the $\times0.6\leftrightarrow\times1.0$ setting. For the $\times0.85\leftrightarrow\times1.0$ setting, we derive the corresponding datasets by center-cropping the images in the $\times0.6\leftrightarrow\times1.0$ datasets. Specifically, the cropping scale is set to $0.6/0.85$ for the UW image and $1$ for the W image. For intermediate images in the training sequences, the cropping scale is linearly increased from $0.6/0.85$ to $1$ along the transition from the UW view to the W view.
\subsection{Experimental Settings}
\label{section:5_1}
\noindent\textbf{Data Construction for ZoomGS v2.}
We first capture multiple dual-camera image pairs to statistically characterize the geometric and color disparities between the UW and W cameras.
For each scene, we capture 15 UW-W image pairs from different viewpoints using a smartphone.
We then use COLMAP~\cite{schonberger2016structure} to estimate the camera poses and intrinsic parameters of the captured images.
For the Xiaomi device, we collect 48 indoor and 30 outdoor image pairs using a Redmi K50 Ultra across diverse scenarios, including \textit{classrooms}, \textit{dining halls}, \textit{shopping malls}, \textit{campuses}, and \textit{companies}.
For the Huawei device, we collect 25 indoor and 25 outdoor pairs using a Huawei Mate 40E.
To construct the 3D scenes, we use 4$,$000 scenes from the DL3DV-10K~\cite{dl3dv} dataset, each of which contains more than 300 views.
For each scene, we randomly sample four dual-camera parameter sets, yielding 16$,$000 synthetic DCSZ zoom sequences per device. After manually removing image sequences with poor 3DGS~\cite{3DGS} reconstruction quality or severe rendering artifacts, 8$,$704 sequences remain for Xiaomi and 7$,$686 for Huawei.

\begin{table*}[t!] 
	\small
	\renewcommand\arraystretch{1}
	\begin{center}
		\caption{
			Quantitative comparisons of VFI models on the Xiaomi test dataset.
            $\dagger$ denotes models fine-tuned on data generated by ZoomGS~\cite{wu2024dual}, and $\ddagger$ denotes models fine-tuned on data generated by ZoomGS v2.
			$\uparrow$ indicates that higher values are better, and $\downarrow$ indicates that lower values are better.
			The best results are marked in \textcolor{red}{red}, while the second-best results are marked in \textcolor{blue}{blue}.
            Models evaluated without fine-tuning produce excessively blurry results, which can artificially improve the color-continuity metrics by suppressing frame-to-frame color variations.
		} 
		\label{tab:real_compare}
		% \vspace{-2mm}
		\scalebox{0.92}{
			\begin{tabular}{p{2cm}<{\centering} p{2.2cm}<{\centering} p{6.9cm}<{\centering}  p{6.9cm}<{\centering}} 
				\toprule
				%\multirow{2}{*}
				\multirow{2}{*}{Finetuning Data} & \multirow{2}{*}{Networks} & \multicolumn{2}{c}{MUSIQ$\uparrow$  / MANIQA$\uparrow$ / TOPIQ$\uparrow$ \ \ DOVER$\uparrow$ / FasterVQA$\uparrow$ \ \ $\mathrm{Smooth}$-$\mathrm{\Delta E}_{mean}$$\downarrow$ / $\mathrm{Smooth}$-$\mathrm{\Delta E}_{max}$$\downarrow$} \\
				& & $(\times0.6 \leftrightarrow \times1.0)$ & $(\times0.85 \leftrightarrow \times1.0)$ \\  
				\midrule
				\multirow{3}{*}{None}& RIFE~\cite{RIFE} & 33.058 / 0.213 / 0.290 \ \ \textcolor{red}{48.26} / \textcolor{blue}{0.666} \ \ \textcolor{blue}{0.142} / \textcolor{blue}{0.445} & 42.830 / 0.246 / 0.321 \ \ 67.65 / 0.688 \ \ \textcolor{red}{0.016} / \textcolor{red}{0.074} \\
				& AMT~\cite{AMT} & \textcolor{red}{43.457} / \textcolor{blue}{0.227} / \textcolor{blue}{0.323} \ \ 44.17 / \textcolor{red}{0.746} \ \ 0.220 / 0.752 & \textcolor{red}{53.093} / \textcolor{red}{0.294} / \textcolor{blue}{0.389} \ \ \textcolor{red}{80.08} / \textcolor{blue}{0.751} \ \ \textcolor{blue}{0.049} / \textcolor{blue}{0.135}
  \\  
				& VFIMamba~\cite{zhang2024vfimamba} & \textcolor{blue}{40.732} / \textcolor{red}{0.240} / \textcolor{red}{0.364} \ \ \textcolor{blue}{44.59} / 0.597 \ \ \textcolor{red}{0.072} / \textcolor{red}{0.426} & \textcolor{blue}{47.434} / \textcolor{blue}{0.272} / \textcolor{red}{0.397} \ \ \textcolor{blue}{69.36} / \textcolor{red}{0.752} \ \ 0.059 / 0.438 \\  
				
				\midrule
                \multirow{5}{2cm}{\centering ZoomGS~\cite{wu2024dual}}& RIFE$^{\dagger}$~\cite{RIFE} & 59.460 / 0.319 / 0.484 \ \ 80.26 / 0.824 \ \ \textcolor{blue}{0.063} / 0.380 & \textcolor{blue}{60.263} / 0.336 / \textcolor{blue}{0.473} \ \ 82.05 / 0.837 \ \ \textcolor{blue}{0.028} / 0.184  \\
				& AMT$^{\dagger}$~\cite{AMT} & 61.007 / 0.329 / 0.478 \ \ 80.98 / 0.833 \ \ 0.162 / 0.513 & 59.704 / 0.329 / 0.460 \ \ 81.77 / 0.838 \ \ 0.136 / 0.387 \\  
				& VFIMamba$^{\dagger}$~\cite{zhang2024vfimamba} & 53.870 / 0.288 / 0.406 \ \ 80.48 / 0.801 \ \ 0.149 / 0.466 & 53.569 / 0.302 / 0.414 \ \ 80.48 / 0.797 \ \ 0.059 / 0.309 \\  
				& ZoomFI-Lite$^{\dagger}$ & \textcolor{blue}{63.139} / \textcolor{red}{0.358} / \textcolor{blue}{0.511} \ \ \textcolor{blue}{81.94} / \textcolor{blue}{0.837} \ \ \textcolor{red}{0.032} / \textcolor{red}{0.167} & 59.988 / \textcolor{blue}{0.345} / \textcolor{blue}{0.473} \ \ \textcolor{blue}{82.39} / \textcolor{blue}{0.840} \ \ \textcolor{red}{0.022} / \textcolor{red}{0.145} \\
                & ZoomFI$^{\dagger}$ & \textcolor{red}{63.689} / \textcolor{blue}{0.354} / \textcolor{red}{0.518} \ \ \textcolor{red}{81.98} / \textcolor{red}{0.838} \ \ \textcolor{red}{0.032} / \textcolor{blue}{0.195} & \textcolor{red}{61.438} / \textcolor{red}{0.348} / \textcolor{red}{0.486} \ \ \textcolor{red}{82.63} / \textcolor{red}{0.843} \ \ \textcolor{red}{0.022} / \textcolor{blue}{0.149} \\ 
                
                \midrule
				
				\multirow{5}{2cm}{\centering ZoomGS v2}& RIFE$^{\ddagger}$~\cite{RIFE} & 65.238 / 0.371 / 0.556 \ \ 81.51 / 0.837 \ \ 0.032 / 0.133 & \textcolor{blue}{63.574} / 0.359 / \textcolor{blue}{0.513} \ \ 82.53 / 0.852 \ \ 0.047 / 0.136  \\
				& AMT$^{\ddagger}$~\cite{AMT} & 64.704 / 0.366 / 0.546 \ \ 82.36 / 0.845 \ \ 0.162 / 0.510  & 62.694 / 0.355 / 0.498 \ \ 82.45 / 0.847 \ \ 0.107 / 0.401 \\  
				& VFIMamba$^{\ddagger}$~\cite{zhang2024vfimamba} & 55.764 / 0.292 / 0.439 \ \ 81.28 / 0.825 \ \ 0.044 / 0.148 & 58.050 / 0.312 / 0.438 \ \ 82.38 / 0.837 \ \ 0.046 / 0.141 \\  
				& ZoomFI-Lite$^{\ddagger}$ & \textcolor{blue}{65.355} / \textcolor{blue}{0.387} / \textcolor{blue}{0.554} \ \ \textcolor{blue}{82.46} / \textcolor{blue}{0.843} \ \ \textcolor{blue}{0.025} / \textcolor{blue}{0.087} & 63.503 / \textcolor{blue}{0.368} / 0.508 \ \ \textcolor{red}{83.08} / \textcolor{red}{0.858} \ \ \textcolor{red}{0.017} / \textcolor{red}{0.075} \\ 
                & ZoomFI$^{\ddagger}$ & \textcolor{red}{66.172} / \textcolor{red}{0.399} / \textcolor{red}{0.567} \ \ \textcolor{red}{82.48} / \textcolor{red}{0.846} \ \ \textcolor{red}{0.023} / \textcolor{red}{0.082} & \textcolor{red}{63.836} / \textcolor{red}{0.376} / \textcolor{red}{0.517} \ \ \textcolor{blue}{83.05} / \textcolor{blue}{0.851} \ \ \textcolor{blue}{0.021} / \textcolor{blue}{0.095} \\ 
				\bottomrule
			\end{tabular}
		}
	\end{center}
	% \vspace{-4mm}
\end{table*}

\begin{table*}[t!] 
	\small
	\renewcommand\arraystretch{1}
	\begin{center}
		\caption{
			Quantitative comparisons of VFI models on the Huawei test dataset.
            $\dagger$ denotes models fine-tuned on data generated by ZoomGS~\cite{wu2024dual}, and $^{\ddagger}$ denotes models fine-tuned on data generated by ZoomGS v2.
			$\uparrow$ indicates that higher values are better, and $\downarrow$ indicates that lower values are better.
			The best results are marked in \textcolor{red}{red}, while the second-best results are marked in \textcolor{blue}{blue}.
            Models evaluated without fine-tuning produce excessively blurry results, which can artificially improve the color-continuity metrics by suppressing frame-to-frame color variations.
		} 
		\label{tab:vbench_compare}
		% \vspace{-2mm}
		\scalebox{0.92}{
			\begin{tabular}{p{2cm}<{\centering} p{2.2cm}<{\centering} p{6.9cm}<{\centering}  p{6.9cm}<{\centering}}
				\toprule
				%\multirow{2}{*}
				\multirow{2}{*}{Finetuning Data} & \multirow{2}{*}{Networks} & \multicolumn{2}{c}{MUSIQ$\uparrow$  / MANIQA$\uparrow$ / TOPIQ$\uparrow$ \ \ DOVER$\uparrow$ / FasterVQA$\uparrow$ \ \ $\mathrm{Smooth}$-$\mathrm{\Delta E}_{mean}$$\downarrow$ / $\mathrm{Smooth}$-$\mathrm{\Delta E}_{max}$$\downarrow$} \\
				& & $(\times0.6 \leftrightarrow \times1.0)$ & $(\times0.85 \leftrightarrow \times1.0)$ \\  
				\midrule
				\multirow{3}{*}{None}& RIFE~\cite{RIFE} & 34.451 / \textcolor{blue}{0.240} / \textcolor{blue}{0.318} \ \ \textcolor{blue}{43.34} / \textcolor{blue}{0.302} \ \ \textcolor{red}{0.079} / \textcolor{red}{0.520} & 39.464 / 0.249 / 0.300 \ \ \textcolor{blue}{62.71} / 0.324 \ \ \textcolor{blue}{0.063} / 0.470 \\
				& AMT~\cite{AMT} & \textcolor{red}{40.899} / 0.233 / 0.311 \ \ \textcolor{red}{73.58} / \textcolor{red}{0.339} \ \ 0.157
 / \textcolor{blue}{0.570} & \textcolor{red}{51.503} / \textcolor{red}{0.287} / \textcolor{blue}{0.367} \ \ \textcolor{red}{79.40} / \textcolor{red}{0.381} \ \ \textcolor{red}{0.060} / \textcolor{red}{0.172}  \\  
				& VFIMamba~\cite{zhang2024vfimamba} & \textcolor{blue}{40.530} / \textcolor{red}{0.263} / \textcolor{red}{0.369} \ \ 40.99 / 0.280 \ \ \textcolor{blue}{0.085} / 0.573 & \textcolor{blue}{44.737} / \textcolor{blue}{0.271} / \textcolor{red}{0.388} \ \ 59.82 / \textcolor{blue}{0.339} \ \ 0.064 / \textcolor{blue}{0.452} \\  
				
				\midrule
                \multirow{5}{2cm}{\centering ZoomGS~\cite{wu2024dual}}& RIFE$^{\dagger}$~\cite{RIFE} & 63.300 / 0.361 / 0.521 \ \ 80.18 / 0.818 \ \ \textcolor{red}{0.037} / 0.225 & \textcolor{blue}{62.767} / \textcolor{blue}{0.353} / \textcolor{blue}{0.476} \ \ 82.45 / \textcolor{red}{0.832} \ \ \textcolor{blue}{0.035} / 0.241  \\
				& AMT$^{\dagger}$~\cite{AMT} & 62.232 / 0.352 / 0.496 \ \ 81.13 / 0.809 \ \ 0.173 / 0.559 & 61.813 / 0.344 / 0.451 \ \ 81.00 / 0.825 \ \ 0.141 / 0.398 \\  
				& VFIMamba$^{\dagger}$~\cite{zhang2024vfimamba} & 47.315 / 0.270 / 0.350 \ \ 79.84 / 0.754 \ \ 0.191 / 0.663 & 51.827 / 0.304 / 0.398 \ \ 80.60 / 0.780 \ \ 0.109 / 0.731 \\  
				& ZoomFI-Lite$^{\dagger}$ & \textcolor{blue}{66.405} / \textcolor{blue}{0.387} / \textcolor{blue}{0.546} \ \ \textcolor{blue}{82.20} / \textcolor{blue}{0.830} \ \ 0.044 / \textcolor{blue}{0.220} & 62.225 / 0.351 / \textcolor{blue}{0.476} \ \ \textcolor{blue}{82.70} / 0.830 \ \ \textcolor{red}{0.034} / \textcolor{red}{0.223} \\
                & ZoomFI$^{\dagger}$ & \textcolor{red}{66.725} / \textcolor{red}{0.395} / \textcolor{red}{0.551} \ \ \textcolor{red}{82.46} / \textcolor{red}{0.836} \ \ \textcolor{blue}{0.039} / \textcolor{red}{0.217} & \textcolor{red}{63.512} / \textcolor{red}{0.362} / \textcolor{red}{0.483} \ \ \textcolor{red}{83.06} / \textcolor{blue}{0.831} \ \ \textcolor{red}{0.034} / \textcolor{blue}{0.237}\\ 
                
                \midrule
				
				\multirow{5}{2cm}{\centering ZoomGS v2}& RIFE$^{\ddagger}$~\cite{RIFE} & 67.918 / 0.402 / \textcolor{blue}{0.578} \ \ 82.37 / 0.830 \ \ 0.027 / 0.086 & \textcolor{blue}{66.287} / \textcolor{blue}{0.378} / \textcolor{blue}{0.516} \ \ 83.08 / 0.851 \ \ 0.041 / 0.114  \\
				& AMT$^{\ddagger}$~\cite{AMT} & 65.806 / 0.374 / 0.546 \ \ 83.10 / \textcolor{blue}{0.838} \ \ 0.178 / 0.536 & 65.699 / 0.366 / 0.500 \ \ 83.60 / 0.855 \ \ 0.124 / 0.415 \\  
				& VFIMamba$^{\ddagger}$~\cite{zhang2024vfimamba} & 53.043 / 0.291 / 0.420 \ \ 80.59 / 0.797 \ \ 0.035 / 0.124 & 59.404 / 0.327 / 0.441 \ \ 82.86 / 0.828 \ \ 0.050 / 0.153 \\  
				& ZoomFI-Lite$^{\ddagger}$ & \textcolor{blue}{68.138} / \textcolor{blue}{0.410} / 0.576 \ \ \textcolor{blue}{83.48} / \textcolor{red}{0.843} \ \ \textcolor{red}{0.023} / \textcolor{red}{0.077} & 66.213 / \textcolor{blue}{0.378} / 0.509 \ \ \textcolor{blue}{83.96} / \textcolor{red}{0.859} \ \ \textcolor{red}{0.019} / \textcolor{red}{0.088} \\ 
                & ZoomFI$^{\ddagger}$ & \textcolor{red}{68.466} / \textcolor{red}{0.422} / \textcolor{red}{0.582} \ \ \textcolor{red}{83.40} / \textcolor{red}{0.843} \ \ \textcolor{blue}{0.025} / \textcolor{blue}{0.083} & \textcolor{red}{66.895} / \textcolor{red}{0.390} / \textcolor{red}{0.521} \ \ \textcolor{red}{84.10} / \textcolor{blue}{0.858} \ \ \textcolor{blue}{0.019} / \textcolor{blue}{0.097} \\ 
				\bottomrule
			\end{tabular}
		}
	\end{center}
	\vspace{-2mm}
\end{table*}

\noindent\textbf{Construction of Test Dataset.}
Obtaining suitable DCSZ test data for reference-based evaluation is challenging. Directly capturing real-world DCSZ sequences would require an array of smartphone cameras with different physical positions and camera parameters, making the setup impractical. 
Synthesizing DCSZ test data with ZoomGS~\cite{wu2024dual} or ZoomGS v2 is also suboptimal. In ZoomGS, inaccuracies in Gaussian reconstruction can produce blurry rendered sequences with visible artifacts, making them unsuitable as ground-truth references. Moreover, the camera poses used by both ZoomGS and ZoomGS v2 are obtained through calibration and may not fully match the poses of real dual-camera systems.
Consequently, evaluating VFI models on these rendered image sequences may not faithfully reflect their performance during real-world zooming.
We therefore evaluate the generated sequences using no-reference image and video quality metrics, which do not require ground-truth references.
Therefore, we directly capture real-world UW/W image pairs and use them to evaluate the generated sequences with no-reference metrics. We assess the VFI results along three dimensions: image quality, video quality, and color continuity, to provide a multidimensional evaluation.
For testing, we additionally collect 100 scenes for Xiaomi and 50 scenes for Huawei. Each scene contains a UW image and its corresponding W image.

\noindent\textbf{Training Configurations.}
We optimize 3DGS representations in ZoomGS v2 using Adam optimizer~\cite{kingma2014adam} for 50k iterations, following the same rendering resolution and optimization schedule of the original 3DGS method.
ZoomFI is first pretrained for 200k iterations on the data generated by ZoomGS v2 without color transformation and then fine-tuned for 100k iterations on color-transformed data.
We augment the training data using random horizontal and vertical flips.
Cosine annealing strategy~\cite{loshchilov2016sgdr} is employed to decrease the learning rate from $0.75\times10^{-4}$ to $0.75\times10^{-6}$ steadily.
All experiments are conducted with PyTorch~\cite{paszke2019pytorch} on two NVIDIA GeForce RTX A6000 GPUs.

\noindent\textbf{Evaluation Configurations.}
Since the test data lack ground-truth references, we employ several no-reference metrics, including MUSIQ~\cite{ke2021musiq}, MANIQA~\cite{yang2022maniqa}, and TOPIQ~\cite{chen2024topiq} for single-image quality assessment, and DOVER~\cite{wu2023exploring}, FasterVQA~\cite{wu2023neighbourhood}
for video quality assessment. 
In addition, we design a protocol for evaluating color continuity. First, we align all frames with the W image using cropping center and optical-flow-based warping~\cite{9722682}, with SEA-RAFT~\cite{searaft} used to estimate the flow. We then compute $\mathrm{\Delta E}$~\cite{mclaren1976xiii} between each aligned frame and the UW image. Next, we fit the color transition curve using a cubic polynomial, which serves as a compromise between modeling nonlinear color transitions and avoiding excessive model complexity. Finally, we compute the mean and maximum absolute differences between the fitted curve and the measured values across frames, denoted by $\mathrm{Smooth}$-$\mathrm{\Delta E}_{mean}$ and $\mathrm{Smooth}$-$\mathrm{\Delta E}_{max}$, respectively.

\begin{figure*}[t!]
	\centering
    \begin{subfigure}{1.0\textwidth}
        \setlength{\abovecaptionskip}{6.mm}
		\ContinuedFloat
        \hfill
		\begin{overpic}[width=0.97\textwidth,grid=False]
			{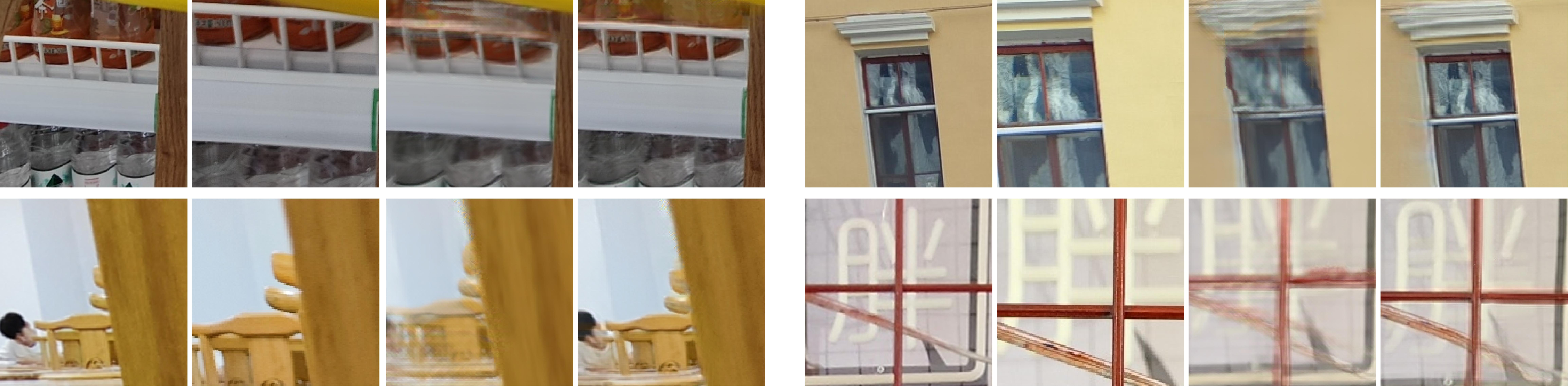}
			\put(13,-8){{\fontsize{8.5}{7}\selectfont UW Image}}
			\put(77,-8){{\fontsize{8.5}{7}\selectfont W Image}}
			\put(131,-8){{\fontsize{8.5}{7}\selectfont ZoomGS~\cite{wu2024dual}}}
            %\put(169,49.5){{\fontsize{6.5}{7}\selectfont (R-ZoomGS)}}
            \put(195,-8){{\fontsize{8.5}{7}\selectfont ZoomGS v2}}
            \put(-10,70){{\fontsize{8.5}{8}\selectfont \rotatebox{90}{$\mathrm{\times0.6 \leftrightarrow \times1.0}$}}}
            \put(-10,4){{\fontsize{8.5}{8}\selectfont \rotatebox{90}{$\mathrm{\times0.85 \leftrightarrow \times1.0}$}}}
            \put(271,-8){{\fontsize{8.5}{7}\selectfont UW Image}}
			\put(334,-8){{\fontsize{8.5}{7}\selectfont W Image}}
			\put(388,-8){{\fontsize{8.5}{7}\selectfont ZoomGS~\cite{wu2024dual}}}
            %\put(169,49.5){{\fontsize{6.5}{7}\selectfont (R-ZoomGS)}}
            \put(450,-8){{\fontsize{8.5}{7}\selectfont ZoomGS v2}}
            %\put(165,-8.5){{\fontsize{6.5}{7}\selectfont (S\&R-ZoomGS)}}
		\end{overpic}
		\vspace{4mm}
	\end{subfigure}
	\caption{Visual comparisons on the Xiaomi test dataset of ZoomFI models trained with data generated by ZoomGS~\cite{wu2024dual} and ZoomGS v2. The top and bottom rows correspond to the $\times0.6\leftrightarrow\times1.0$ and $\times0.85\leftrightarrow\times1.0$ settings, respectively.} 
	\label{fig:real_vis_comp_xiaomi06_1}
    % \vspace{-5mm}
\end{figure*}
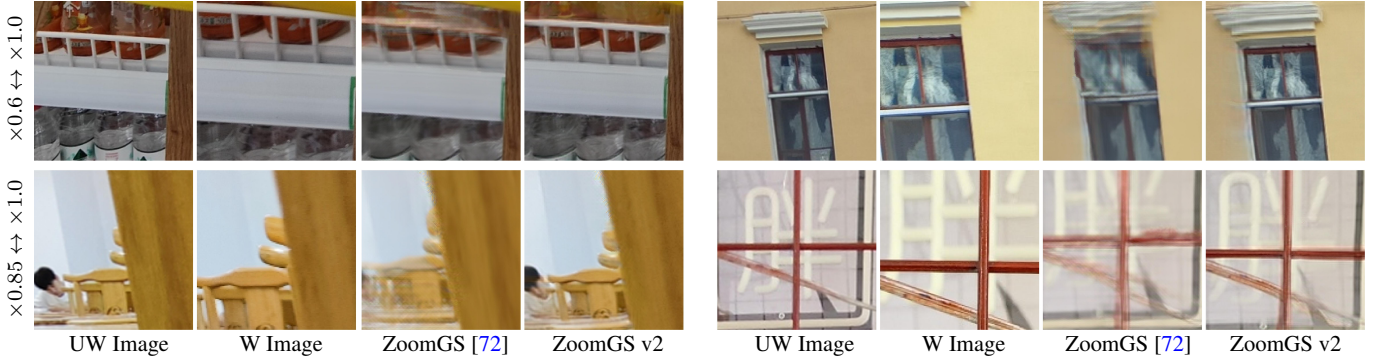

\begin{figure*}[t!]
	\centering
    \begin{subfigure}{1.0\textwidth}
        \setlength{\abovecaptionskip}{6.mm}
		\ContinuedFloat
        \hfill
		\begin{overpic}[width=0.97\textwidth,grid=False]
			{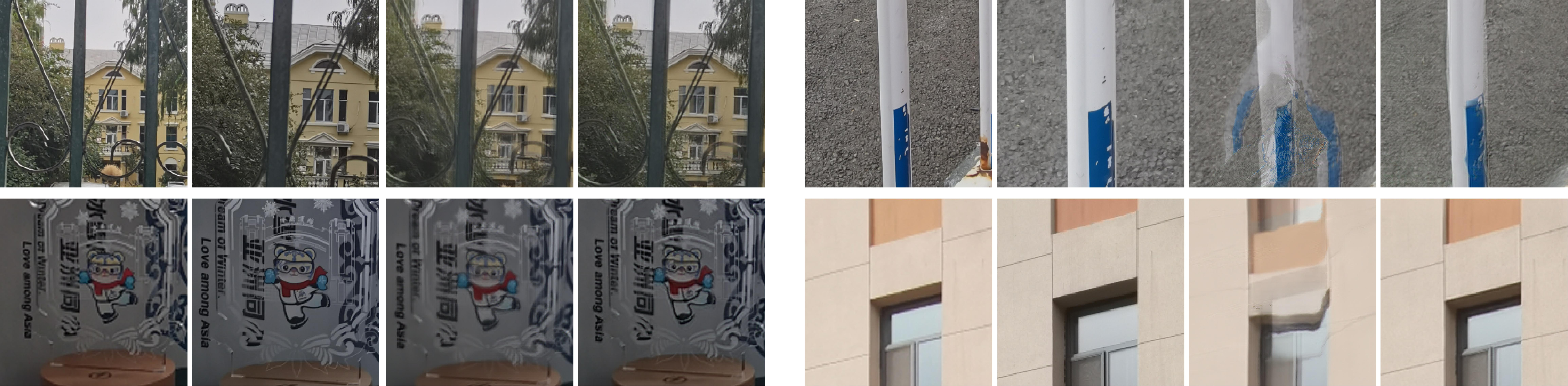}
			\put(13,-8){{\fontsize{8.5}{7}\selectfont UW Image}}
			\put(77,-8){{\fontsize{8.5}{7}\selectfont W Image}}
			\put(131,-8){{\fontsize{8.5}{7}\selectfont ZoomGS~\cite{wu2024dual}}}
            %\put(169,49.5){{\fontsize{6.5}{7}\selectfont (R-ZoomGS)}}
            \put(195,-8){{\fontsize{8.5}{7}\selectfont ZoomGS v2}}
            \put(-10,70){{\fontsize{8.5}{8}\selectfont \rotatebox{90}{$\mathrm{\times0.6 \leftrightarrow \times1.0}$}}}
            \put(-10,4){{\fontsize{8.5}{8}\selectfont \rotatebox{90}{$\mathrm{\times0.85 \leftrightarrow \times1.0}$}}}
            \put(271,-8){{\fontsize{8.5}{7}\selectfont UW Image}}
			\put(334,-8){{\fontsize{8.5}{7}\selectfont W Image}}
			\put(388,-8){{\fontsize{8.5}{7}\selectfont ZoomGS~\cite{wu2024dual}}}
            %\put(169,49.5){{\fontsize{6.5}{7}\selectfont (R-ZoomGS)}}
            \put(450,-8){{\fontsize{8.5}{7}\selectfont ZoomGS v2}}
            %\put(165,-8.5){{\fontsize{6.5}{7}\selectfont (S\&R-ZoomGS)}}
		\end{overpic}
		\vspace{4mm}
	\end{subfigure}
	\caption{Visual comparisons on the Huawei test dataset of ZoomFI models trained with data generated by ZoomGS~\cite{wu2024dual} and ZoomGS v2. The top and bottom rows correspond to the $\times0.6\leftrightarrow\times1.0$ and $\times0.85\leftrightarrow\times1.0$ settings, respectively.} 
	\label{fig:real_vis_comp_huawei06_1}
    \vspace{0mm}
\end{figure*}
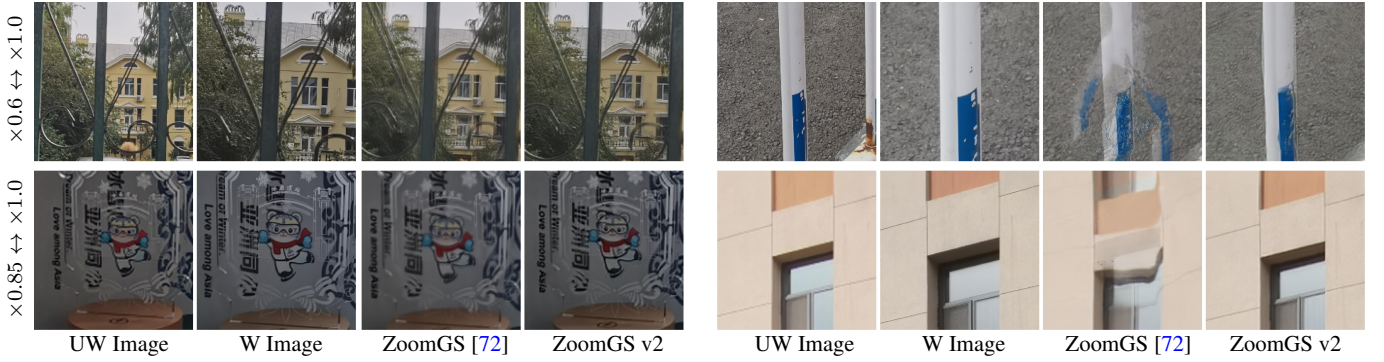

\subsection{Comparison with State-of-the-Art Methods}
\label{section:5_2}
\noindent\textbf{Comparison Configurations.}
We compare our approach with three representative VFI methods (\ie, RIFE~\cite{RIFE}, AMT~\cite{AMT}, and VFIMamba~\cite{zhang2024vfimamba}) to evaluate the effectiveness of ZoomGS v2 and ZoomFI.
All VFI models are trained under the same configurations to ensure a fair comparison.

\noindent\textbf{Quantitative and Qualitative Results.}
As shown in Table~\ref{tab:real_compare} and Table~\ref{tab:vbench_compare},
models fine-tuned on data generated by ZoomGS v2 consistently outperform those fine-tuned on ZoomGS~\cite{wu2024dual} and their non-fine-tuned counterparts on most evaluation metrics, with particularly notable improvements in perceptual quality. These results support the effectiveness of ZoomGS v2 in improving VFI performance in real-world DCSZ scenarios. 
Non-fine-tuned models produce overly blurry results that suppress high-frequency color details. This can artificially reduce frame-to-frame fluctuations in $\mathrm{\Delta E}$~\cite{mclaren1976xiii}, making the color continuity metrics appear better than the actual visual quality would suggest.
Visual comparisons in Fig.~\ref{fig:real_vis_comp_xiaomi06_1} and Fig.~\ref{fig:real_vis_comp_huawei06_1} further support these findings, showing that models fine-tuned on ZoomGS v2 data preserve more photorealistic details and exhibit fewer visual artifacts.
Across the evaluated settings, ZoomFI achieves the best performance on most evaluation metrics among the compared methods, as illustrated in Fig.~\ref{fig:syn_vis_comp_xiaomi} and Fig.~\ref{fig:real_vis_comp_xiaomi}.

\noindent\textbf{Efficiency of VFI Models.}
As shown in Table~\ref{tab:flop}, under identical hardware and implementation settings, ZoomFI has a shorter inference time than VFIMamba and AMT, whereas ZoomFI-Lite achieves the lowest inference time and FLOPs count among all compared methods. These results indicate that ZoomFI balances visual quality and inference speed, whereas ZoomFI-Lite offers the lowest computational cost and maintains competitive generation quality relative to the other methods, except ZoomFI. Both models are promising candidates for mobile deployment. All evaluations are performed at a resolution of $1632\times1216$.

\begin{figure*}[t!]
	\centering
	\begin{subfigure}{0.96\textwidth}
        \hfill
        \setlength{\abovecaptionskip}{5.mm}
		\ContinuedFloat
		\begin{overpic}[width=1.0\textwidth,grid=False]
			{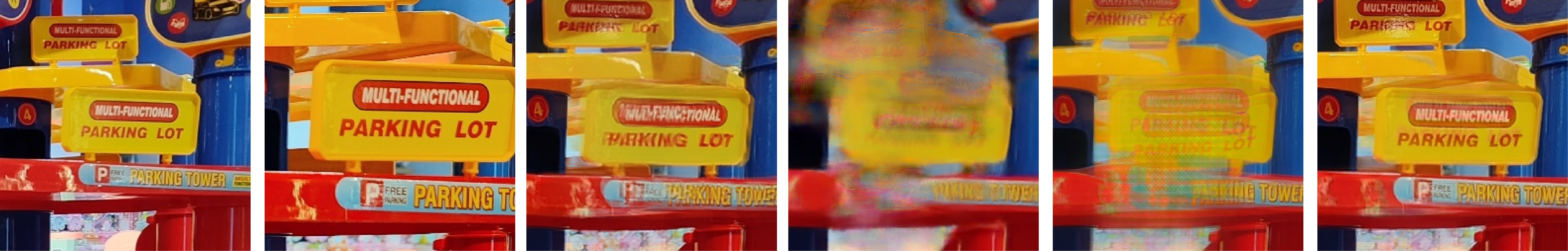}
		\end{overpic}
		\vspace{0.1mm}
	\end{subfigure}
    \begin{subfigure}{0.96\textwidth}
        \hfill
        \setlength{\abovecaptionskip}{5.mm}
		\ContinuedFloat
		\begin{overpic}[width=1.0\textwidth,grid=False]
			{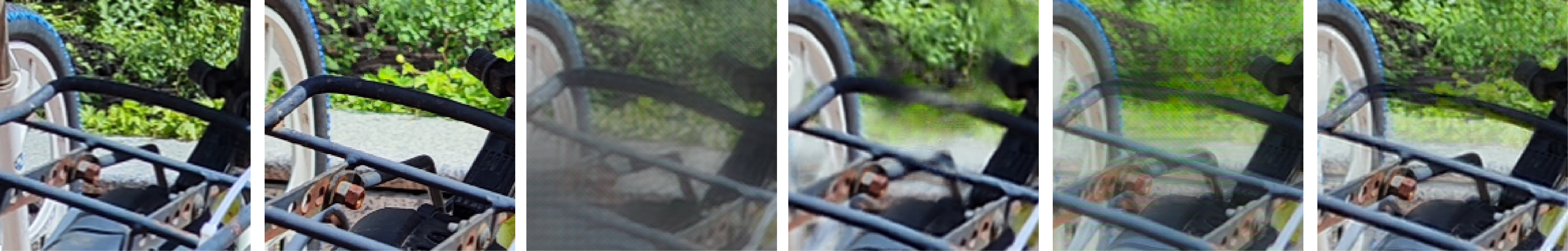}
			\put(25,-9){{\fontsize{8.5}{7}\selectfont UW Image}}
			\put(108,-9){{\fontsize{8.5}{7}\selectfont W Image}}
			\put(190,-9){{\fontsize{8.5}{7}\selectfont AMT$^{\ddagger}$~\cite{AMT}}}
			\put(262,-9){{\fontsize{8.5}{7}\selectfont VFIMamba$^{\ddagger}$~\cite{zhang2024vfimamba}}}
			\put(354,-9){{\fontsize{8.5}{7}\selectfont RIFE$^{\ddagger}$~\cite{RIFE}}}
			\put(440,-9){{\fontsize{8.5}{7}\selectfont ZoomFI$^{\ddagger}$}}
            \put(-10,105){{\fontsize{8.5}{7}\selectfont \rotatebox{90}{$\mathrm{\times0.6 \leftrightarrow \times1.0}$}}}
            \put(-10,10){{\fontsize{8.5}{7}\selectfont \rotatebox{90}{$\mathrm{\times0.85 \leftrightarrow \times1.0}$}}}
		\end{overpic}
		\vspace{-2mm}
	\end{subfigure}
	
	\caption{Visual comparisons of different VFI models on the Xiaomi test dataset. The top and bottom rows correspond to the $\times0.6\leftrightarrow\times1.0$ and $\times0.85\leftrightarrow\times1.0$ settings, respectively.
	} 
	\label{fig:syn_vis_comp_xiaomi}
	\vspace{0mm}
\end{figure*}
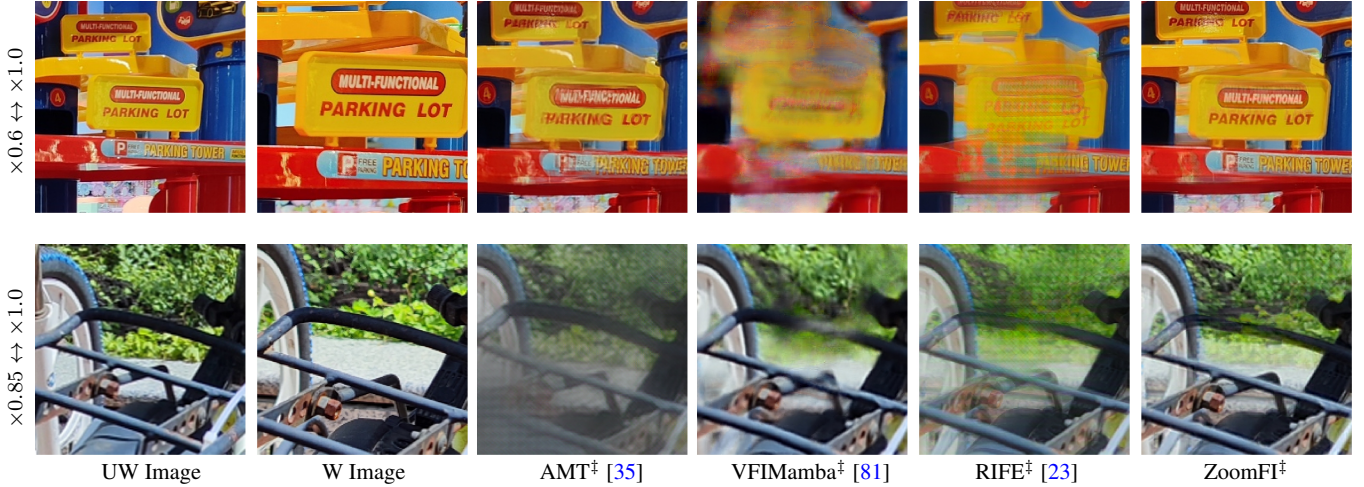

\begin{figure*}[t!]
	\centering
	\begin{subfigure}{0.96\textwidth}
		\hfill
        \setlength{\abovecaptionskip}{5.mm}
        \ContinuedFloat
		\begin{overpic}[width=1.0\textwidth,grid=False]
			{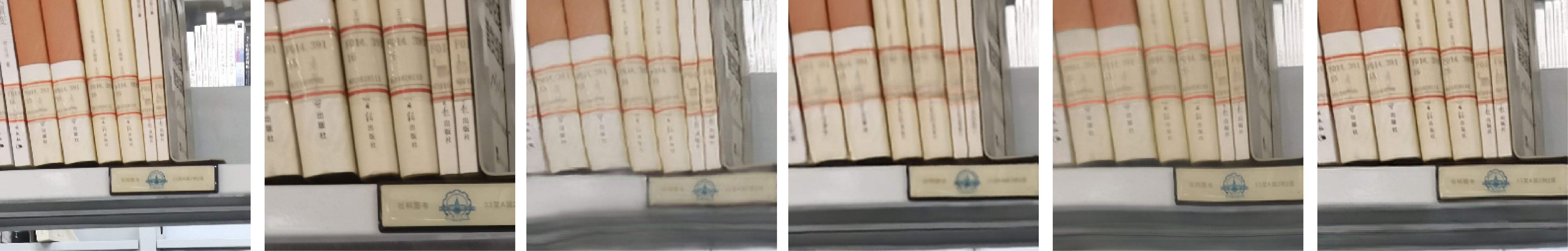}
		\end{overpic}
		\vspace{0.1mm}
	\end{subfigure}
    \begin{subfigure}{0.96\textwidth}
		\hfill
        \setlength{\abovecaptionskip}{5.mm}
        \ContinuedFloat
		\begin{overpic}[width=1.0\textwidth,grid=False]
			{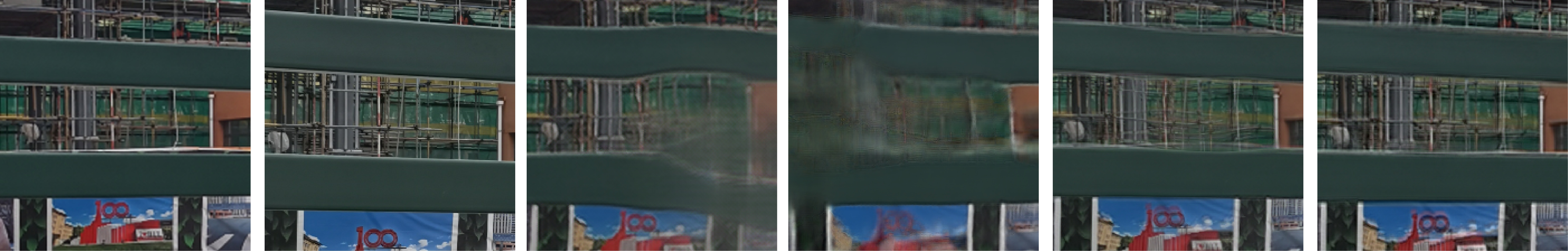}
			\put(25,-9){{\fontsize{8.5}{7}\selectfont UW Image}}
			\put(108,-9){{\fontsize{8.5}{7}\selectfont W Image}}
			\put(190,-9){{\fontsize{8.5}{7}\selectfont AMT$^{\ddagger}$~\cite{AMT}}}
			\put(262,-9){{\fontsize{8.5}{7}\selectfont VFIMamba$^{\ddagger}$~\cite{zhang2024vfimamba}}}
			\put(354,-9){{\fontsize{8.5}{7}\selectfont RIFE$^{\ddagger}$~\cite{RIFE}}}
			\put(440,-9){{\fontsize{8.5}{7}\selectfont ZoomFI$^{\ddagger}$}}
            \put(-10,105){{\fontsize{8.5}{8}\selectfont \rotatebox{90}{$\mathrm{\times0.6 \leftrightarrow \times1.0}$}}}
            \put(-10,10){{\fontsize{8.5}{8}\selectfont \rotatebox{90}{$\mathrm{\times0.85 \leftrightarrow \times1.0}$}}}
		\end{overpic}
		\vspace{-2mm}
	\end{subfigure}
	
	\caption{Visual comparisons of different VFI models on the Huawei test dataset. The top and bottom rows correspond to the $\times0.6\leftrightarrow\times1.0$ and $\times0.85\leftrightarrow\times1.0$ settings, respectively.
	} 
	\label{fig:real_vis_comp_xiaomi}
	\vspace{0mm}
\end{figure*}

\begin{table}[t!] 
	\small
	\renewcommand\arraystretch{1}
	\begin{center}
		\caption{
            Comparison of FLOPs, parameters, and inference time of different VFI models at a test resolution of $1632\times1216$.
		} 
		\label{tab:flop}
		% \vspace{-2mm}
        \setlength{\tabcolsep}{3pt}
		% \scalebox{0.75}{
        \scalebox{0.97}{
			\begin{tabular}{p{2.3cm}<{\centering} p{1.8cm}<{\centering} p{2.2cm}<{\centering} p{1.8cm}<{\centering}} 
				\toprule
				%\multirow{2}{*}
				Networks & FLOPs (G) & Parameters (M) & Time (ms) \\  
				\midrule
				
				RIFE~\cite{RIFE} & 708.3 & 10.1 & 48.7 \\
                AMT~\cite{AMT} & 540.3 & 3.0 & 183.2 \\
				VFIMamba~\cite{zhang2024vfimamba} & 1792.9 & 16.1 & 159.7 \\
                ZoomFI-Lite & 422.7 & 12.5  & 40.9 \\
                ZoomFI & 724.2 & 12.5 & 52.1 \\
				\bottomrule
			\end{tabular}
		}
	\end{center}
	\vspace{-2mm}
\end{table}

\subsection{Ablation Study}
We conduct ablation studies to evaluate the two main components of our approach: the ZoomGS v2 data generation pipeline and ZoomFI. We also investigate how the amount of data generated by ZoomGS v2 affects fine-tuning.
All ablation experiments are conducted on Xiaomi smartphone device and $\times 0.6 \leftrightarrow \times 1.0$ setting.

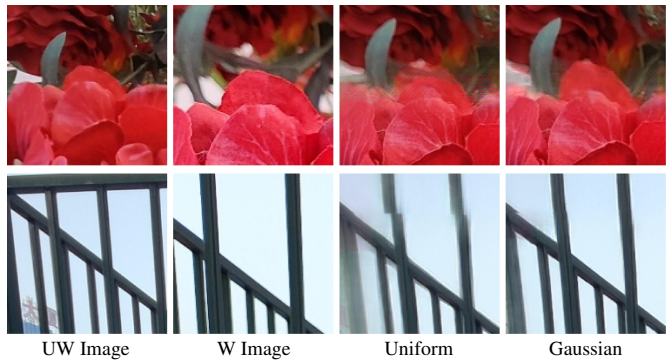
\begin{figure}[t!]
	\centering
    \begin{subfigure}{0.48\textwidth}
        \setlength{\abovecaptionskip}{6.mm}
		\ContinuedFloat
		\begin{overpic}[width=1.0\textwidth,grid=False]
			{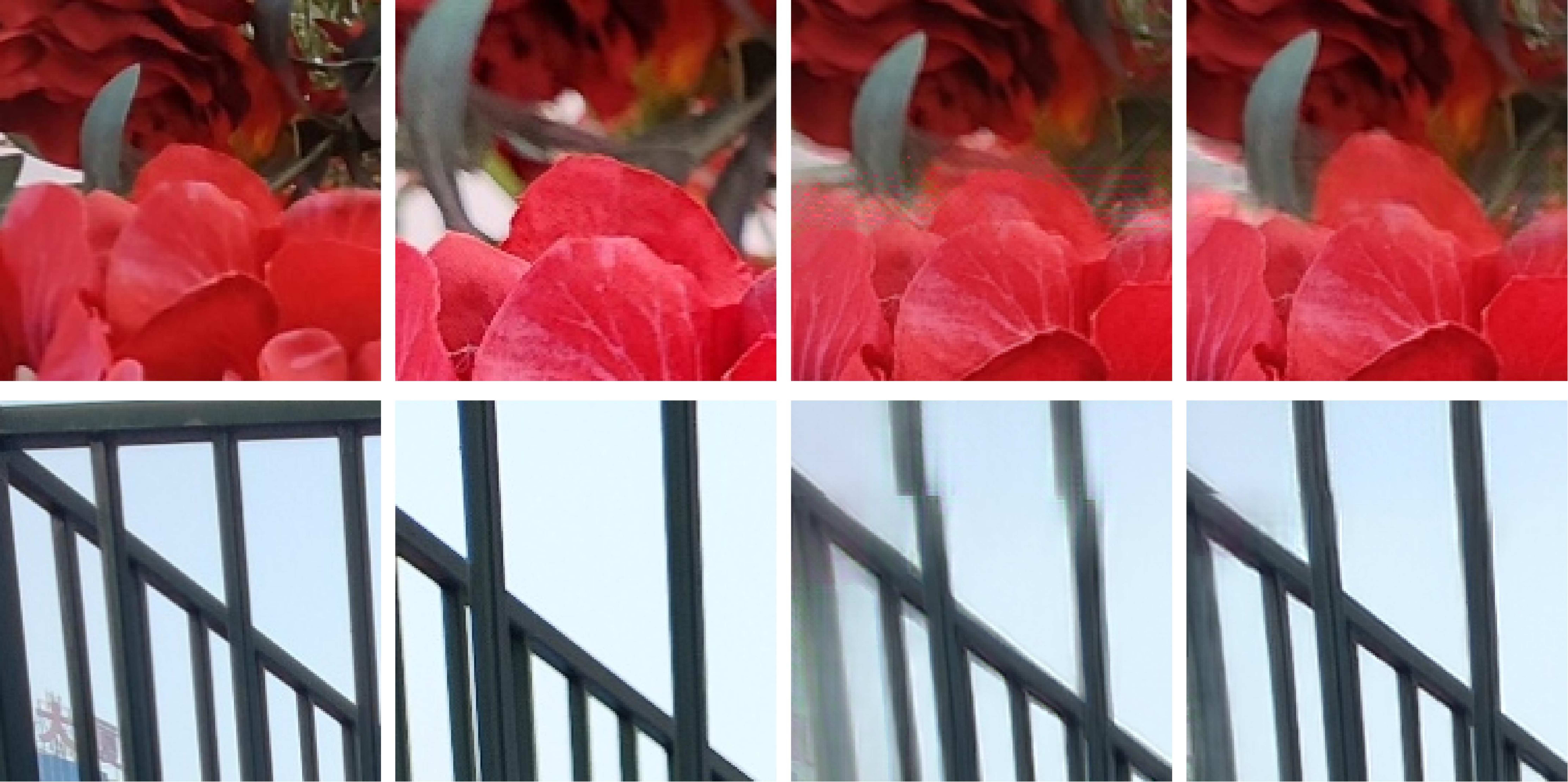}
			\put(13,-8){{\fontsize{7.5}{7}\selectfont UW Image}}
			\put(79,-8){{\fontsize{7.5}{7}\selectfont W Image}}
			\put(142,-8){{\fontsize{7.5}{7}\selectfont Uniform}}
            %\put(169,49.5){{\fontsize{6.5}{7}\selectfont (R-ZoomGS)}}
            \put(204,-8){{\fontsize{7.5}{7}\selectfont Gaussian}}
            %\put(165,-8.5){{\fontsize{6.5}{7}\selectfont (S\&R-ZoomGS)}}
		\end{overpic}
		\vspace{0mm}
	\end{subfigure}
	\caption{Visual comparison of different sampling strategies for $\Delta\mathbf{V}$. Gaussian sampling produces fewer misalignment artifacts and better visual results than Uniform sampling.} 
	\label{fig:position}
    \vspace{-2mm}
\end{figure}

\noindent\textbf{Effect of Sampling Strategy of $\Delta \mathbf{V}$ in ZoomGS v2.}
We model the relative translation offsets $\Delta \mathbf{V}$ with a multivariate Gaussian distribution (Sec.~\ref{section:4_0_3}) and sample from it to generate synthetic zoom data.
For a fair comparison, Gaussian and uniform sampling use the same number of samples and identical sampling ranges, with the bounds of the uniform distribution are set to the minimum and maximum values observed in the real dual-camera dataset. All other training configurations follow Sec.~\ref{section:5_1}.
Compared with uniform sampling, Gaussian sampling improves all image and video quality metrics and one color continuity metric, supporting the effectiveness of the multivariate Gaussian sampling, as shown in Table~\ref{tab:dl3dv_sample_method}. The visual comparisons in Fig.~\ref{fig:position} are consistent with these results: Gaussian sampling produces substantially fewer misalignment artifacts than Uniform sampling.
This improvement may be attributed to the closer match between Gaussian sampling and the empirical distribution of real dual-camera translations. As shown in Fig.~\ref{fig_trans}, the measured relative translation offsets are concentrated in a limited region rather than distributed uniformly. Uniform sampling therefore produces more atypical camera configurations, increasing the distribution gap between the synthetic training data and real DCSZ data.

\begin{table}[t!] 
	\small
	\renewcommand\arraystretch{1}
	\begin{center}
		\caption{
			Ablation studies of the $\Delta \mathbf{V}$ sampling strategy in ZoomGS v2.
		} 
		\label{tab:dl3dv_sample_method}
		% \vspace{-2mm}
        \setlength{\tabcolsep}{3pt}
		\scalebox{0.77}{
			\begin{tabular}{p{1.7cm}<{\centering} p{3.4cm}<{\centering} p{2.0cm}<{\centering} p{3.0cm}<{\centering}} 
				\toprule
				%\multirow{2}{*}
				$\Delta \mathbf{V}$ sample & MUSIQ$\uparrow$ / MANIQA$\uparrow$ / & DOVER$\uparrow$ / & $\mathrm{Smooth}$-$\mathrm{\Delta E}_{mean}$$\downarrow$ / \\  
                manner & TOPIQ$\uparrow$ & FasterVQA$\uparrow$ &$\mathrm{Smooth}$-$\mathrm{\Delta E}_{max}$$\downarrow$ \\
				\midrule
                Uniform & 65.692 / 0.397 / 0.561 & 82.12 / 0.841 & \pmb{0.021} / 0.096   \\
				Gaussian & 66.172 / 0.399 / \pmb{0.567} & \pmb{82.48} / \pmb{0.846} & 0.023 / \pmb{0.082} \\
				\bottomrule
			\end{tabular}
		}
	\end{center}
	\vspace{-2mm}
\end{table}

\begin{figure}[t!] % [t]
	\centering
    \hfill
	\includegraphics[width=3.3in]{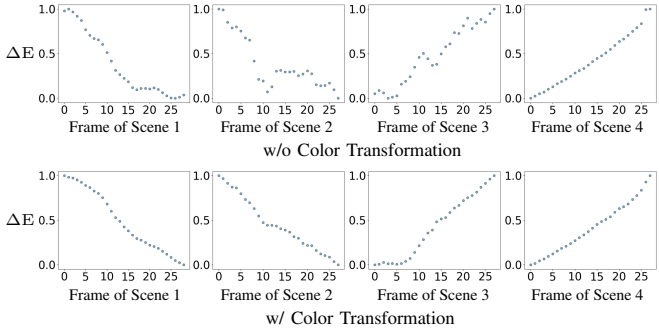}%
    \put(-249,87){{\fontsize{6}{5}\selectfont $\mathrm{\Delta E}$}}
    \put(-249,24){{\fontsize{6}{5}\selectfont $\mathrm{\Delta E}$}}
    \put(-225,59){{\fontsize{6}{5}\selectfont Frame of Scene 1}}
    \put(-168,59){{\fontsize{6}{5}\selectfont Frame of Scene 2}}
    \put(-225,-4){{\fontsize{6}{5}\selectfont Frame of Scene 1}}
    \put(-168,-4){{\fontsize{6}{5}\selectfont Frame of Scene 2}}
    \put(-109,59){{\fontsize{6}{5}\selectfont Frame of Scene 3}}
    \put(-51,59){{\fontsize{6}{5}\selectfont Frame of Scene 4}}
    \put(-109,-4){{\fontsize{6}{5}\selectfont Frame of Scene 3}}
    \put(-51,-4){{\fontsize{6}{5}\selectfont Frame of Scene 4}}
    \put(-152,50){{\fontsize{7}{6}\selectfont w/o Color Transformation}}
    \put(-151,-13){{\fontsize{7}{6}\selectfont w/ Color Transformation}}
	\vspace{0mm}
    \caption{Color-difference trajectories of sequences generated by ZoomFI models trained with and without color transformation. For each sequence, we compute $\mathrm{\Delta E}$~\cite{mclaren1976xiii} between each generated frame and its corresponding UW image. Four representative scenes are shown as examples, with each column corresponding to one scene. The $\mathrm{\Delta E}$ values are normalized within each plot.}
    % \vspace{-2mm}
	\label{color}
\end{figure}

\noindent\textbf{Effect of Color Transformation in ZoomGS v2.}
We evaluate the effectiveness of color transformation and its components. The training configurations follow Sec.~\ref{section:5_1}, except that the fine-tuning stage with color-transformed data is omitted or changed. 
Table~\ref{tab:color} reports the ablation results for color transformation. 
Adding color transformation improves all video-quality and color-continuity metrics over the baseline without color transformation, indicating that reducing the photometric gap improves VFI performance. Moreover, the full color transformation framework outperforms variants that omit either gamma correction or the simplified linear-mapping module on all seven metrics, demonstrating the contribution of both components.
Visual results further support these findings. As illustrated in Fig.~\ref{color}, we calculate $\mathrm{\Delta E}$~\cite{mclaren1976xiii} between each image sequences generated by ZoomFI trained on data with and without color transformation. For each sequence, we compute $\mathrm{\Delta E}$ between each generated frame and its corresponding UW images. The color-transformed sequences exhibit smoother variations in $\mathrm{\Delta E}$ than the untransformed sequences, indicating the effectiveness of color transformation in improving color continuity. 
Furthermore, color-transformed UW images are more color-consistent with the W images and exhibit less chromatic aberration (CA) than the uncorrected images, as shown in Fig.~\ref{color_transform}. These results further demonstrate the effectiveness of the color transformation module.

\begin{table}[t!] 
	\small
	\renewcommand\arraystretch{1}
	\begin{center}
		\caption{
			Ablation study of color-transformation module and its gamma-correction and linear-mapping components.
		} 
		\label{tab:color}
		% \vspace{-2mm}
        \setlength{\tabcolsep}{2pt}
		\scalebox{0.77}{
			\begin{tabular}{p{1.0cm}<{\centering} p{1.8cm}<{\centering} p{3.2cm}<{\centering} p{1.8cm}<{\centering} p{2.8cm}<{\centering}} 
				\toprule
				%\multirow{2}{*}
				Gamma & Linear Map & MUSIQ$\uparrow$ / MANIQA$\uparrow$ / & DOVER$\uparrow$ / & $\mathrm{Smooth}$-$\mathrm{\Delta E}_{mean}$$\downarrow$ / \\  
                Params & Params & / TOPIQ$\uparrow$ & FasterVQA$\uparrow$ &$\mathrm{Smooth}$-$\mathrm{\Delta E}_{max}$$\downarrow$ \\
				\midrule
                $\times$ & $\times$ & \pmb{66.198} / \pmb{0.400} / 0.566 & 82.36 / 0.845 & 0.029 / 0.090 \\
                $\times$ & $\checkmark$ & 66.028 / 0.397 / 0.565 & 82.39 / 0.844 & 0.026 / 0.084 \\
                $\checkmark$ & $\times$ & 63.022 / 0.370 / 0.520 & 81.76 / 0.829 & 0.189 / 0.441 \\
				$\checkmark$ & $\checkmark$ & 66.172 / 0.399 / \pmb{0.567} & \pmb{82.48} / \pmb{0.846} & \pmb{0.023} / \pmb{0.082}  \\
				\bottomrule
			\end{tabular}
		}
	\end{center}
    \vspace{-2mm}
	% \vspace{-8mm}
\end{table}

\noindent\textbf{Effect of Training Data Scale from ZoomGS v2.}
To evaluate the contribution of ZoomGS v2, we increase the proportion of scenes used for training from 1\% to 100\% according to an exponential schedule. The training configurations follow Sec.~\ref{section:5_1}, except that the number of training iterations is scaled in proportion to the dataset size so that each training sample is accessed approximately the same number of times.
As summarized in Table~\ref{tab:A}, the VFI model achieves its best overall performance when trained on the full ZoomGS v2 dataset. Overall, performance improves as the amount of synthetic training data increases, demonstrating the benefits of scaling up the training dataset.
Moreover, when trained on only $1.0\%$ of the ZoomGS v2 scenes, the VFI model performs comparably to or better than the model trained on ZoomGS data, as reported in Table~\ref{tab:real_compare}. The number of scenes in this subset is comparable to that in the ZoomGS dataset, suggesting that ZoomGS v2 provides more effective training data even at a similar scene count.

\begin{figure}[t!] % [t]
	\centering
	\includegraphics[width=3.5in]{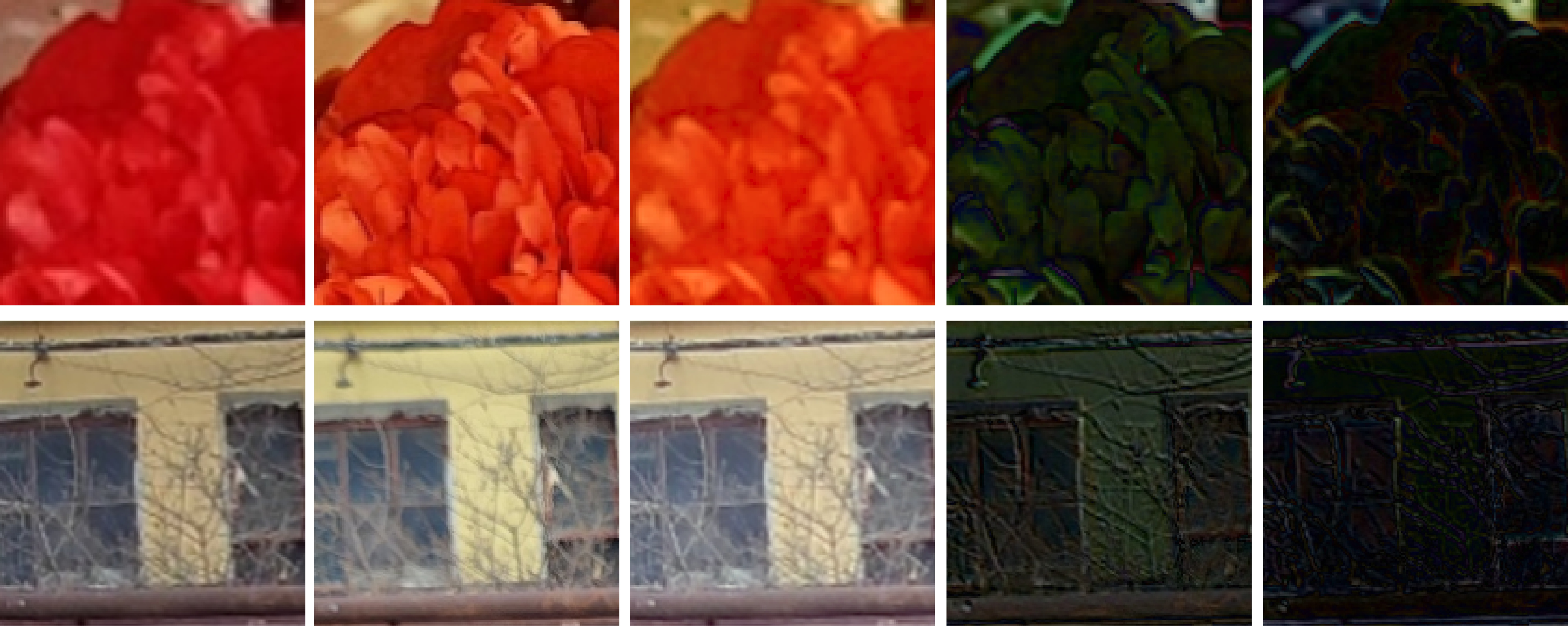}%
    \put(-242,-8){{\fontsize{7.5}{5}\selectfont UW Image}}
	\put(-191,-8){{\fontsize{7.5}{5}\selectfont W Image}}
    \put(-137,-8){{\fontsize{7.5}{5}\selectfont UW$_{\mathrm{T}}$}}
    \put(-99,-8){{\fontsize{7.5}{5}\selectfont CA of UW\&W}}
    \put(-49,-8){{\fontsize{7.5}{5}\selectfont CA of UW$_{\mathrm{T}}$\&W}}
    \vspace{1mm}
	\caption{Visual comparison of the UW images before and after color transformation.}
	\label{color_transform}
    % \vspace{-4mm}
\end{figure}
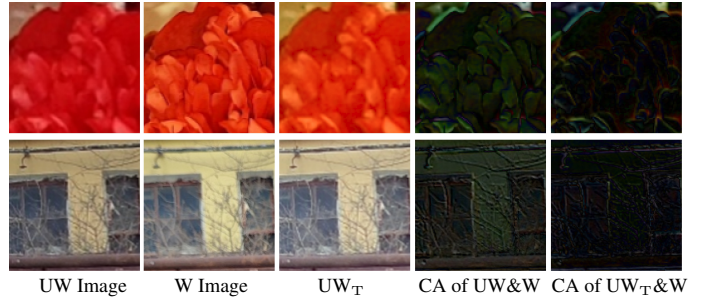

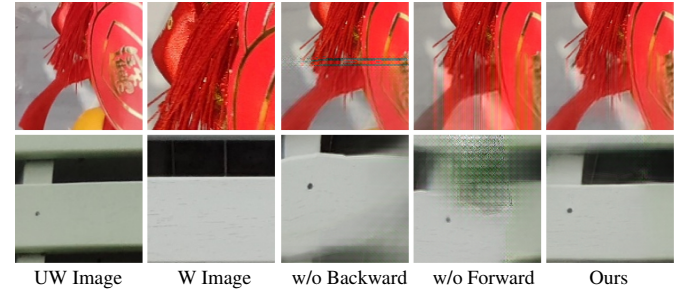
\begin{figure}[t!]
	\centering
    \begin{subfigure}{0.48\textwidth}
        \setlength{\abovecaptionskip}{6.mm}
		\ContinuedFloat
		\begin{overpic}[width=1.0\textwidth,grid=False]
			{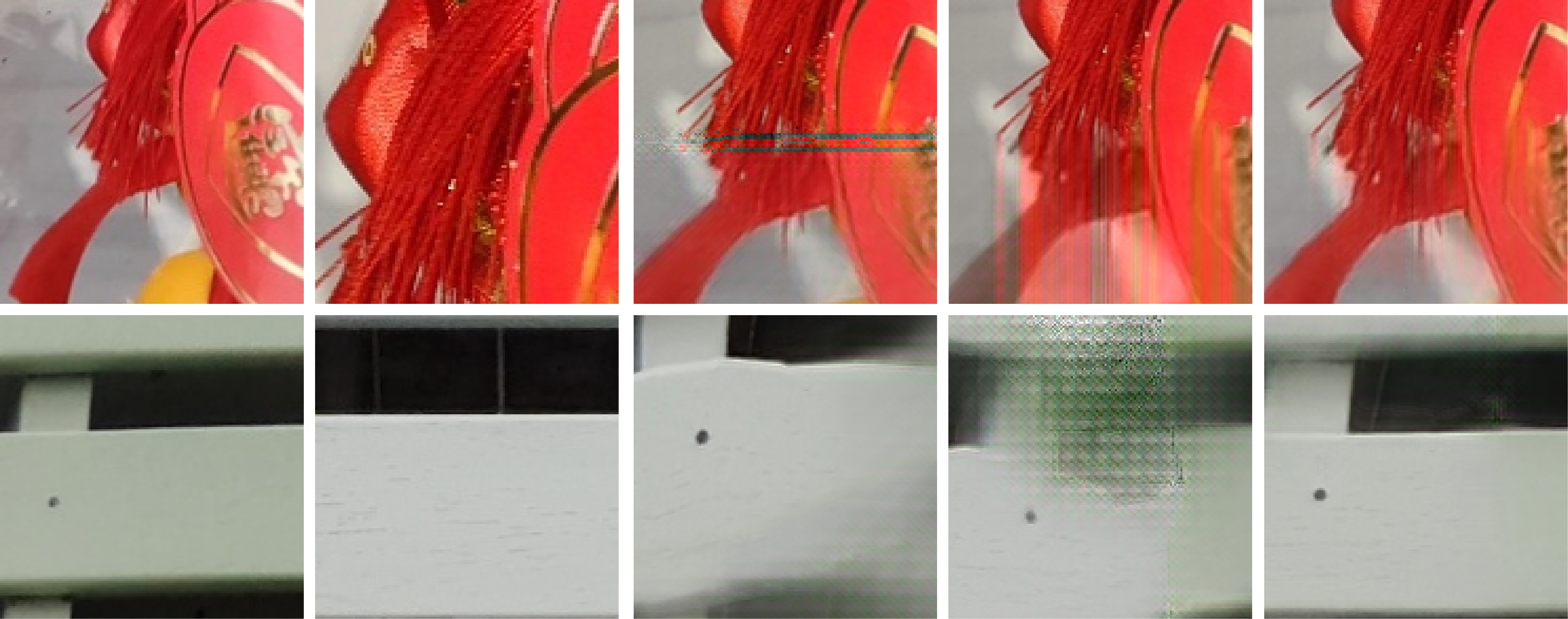}
			\put(7,-8){{\fontsize{7.5}{7}\selectfont UW Image}}
			\put(61,-8){{\fontsize{7.5}{7}\selectfont W Image}}
			\put(104,-8){{\fontsize{7.5}{7}\selectfont w/o Backward}}
            \put(157,-8){{\fontsize{7.5}{7}\selectfont w/o Forward}}
            %\put(169,49.5){{\fontsize{6.5}{7}\selectfont (R-ZoomGS)}}
            \put(216,-8){{\fontsize{7.5}{7}\selectfont Ours}}
            %\put(165,-8.5){{\fontsize{6.5}{7}\selectfont (S\&R-ZoomGS)}}
		\end{overpic}
		\vspace{0mm}
	\end{subfigure}
	\caption{Visual comparisons of ZoomFI with different combinations of forward and backward warping. Combining both warping directions yields better visual results than using either direction alone.} 
	\label{fig:zoomfi}
    % \vspace{-4mm}
    % \vspace{-4mm}
\end{figure}

\begin{table}[t!] 
	\small
	\renewcommand\arraystretch{1}
	\begin{center}
		\caption{
			% Ablation studies on add or do not add forward flow in ZoomFI model. The ablation experiment was conducted on a XIAOMI phone model with zoom factor of UW image set to 0.6.
			% % 
            Effect of training data scale on VFI performance using data generated by ZoomGS v2.
		} 
		\label{tab:A}
		% \vspace{-2mm}
        \setlength{\tabcolsep}{1pt}
		\scalebox{0.82}{
			\begin{tabular}{p{1.8cm}<{\centering} p{3.2cm}<{\centering} p{2.3cm}<{\centering} p{2.9cm}<{\centering}}  
				\toprule
				%\multirow{2}{*}
				Percentage &  MUSIQ$\uparrow$ / MANIQA$\uparrow$ / & DOVER$\uparrow$ / & $\mathrm{Smooth}$-$\mathrm{\Delta E}_{mean}$$\downarrow$ / \\
                of Data & TOPIQ$\uparrow$ & FasterVQA$\uparrow$ & $\mathrm{Smooth}$-$\mathrm{\Delta E}_{max}$$\downarrow$ \\
				\midrule
                1.0\% & 64.324 / 0.385 / 0.539 & 82.02 / 0.837 & 0.036 / 0.117 \\
                10\% & 65.174 / 0.393 / 0.551 & 82.06 / 0.841 & 0.034 / 0.110 \\
                \midrule
                100\% & \pmb{66.172} / \pmb{0.399} / \pmb{0.567} & \pmb{82.48} / \pmb{0.846} & \pmb{0.023} / \pmb{0.082} \\
				\bottomrule
			\end{tabular}
		}
	\end{center}
	% \vspace{-2mm}
\end{table}

\begin{table}[t!] 
	\small
	\renewcommand\arraystretch{1}
	\begin{center}
		\caption{
			% Ablation studies on add or do not add forward flow in ZoomFI model. The ablation experiment was conducted on a XIAOMI phone model with zoom factor of UW image set to 0.6.
			% % 
            Ablation study of backward and forward warping in ZoomFI.
		} 
		\label{tab:forward_flow}
		% \vspace{-2mm}
        \setlength{\tabcolsep}{3pt}
		\scalebox{0.76}{
			\begin{tabular}{p{1.3cm}<{\centering} p{1.3cm}<{\centering} p{3.2cm}<{\centering} p{1.8cm}<{\centering} p{2.8cm}<{\centering}} 
				\toprule
				%\multirow{2}{*}
				backward & forward & MUSIQ$\uparrow$ / MANIQA$\uparrow$ / & DOVER$\uparrow$ / & $\mathrm{Smooth}$-$\mathrm{\Delta E}_{mean}$$\downarrow$ / \\  
                warp & warp & TOPIQ$\uparrow$ & FasterVQA$\uparrow$ & $\mathrm{Smooth}$-$\mathrm{\Delta E}_{max}$$\downarrow$ \\
				\midrule
				
				$\checkmark$ & $\times$ & 65.917 / 0.391 / 0.552 & 82.39 / 0.847 & 0.032 / 0.096  \\
				$\times$ & $\checkmark$ & 65.229 / 0.384 / 0.553 & 82.05 / 0.839 & 0.045 / 0.129  \\
                $\checkmark$ & $\checkmark$ & \pmb{66.172} / \pmb{0.399} / \pmb{0.567} & \pmb{82.48} / \pmb{0.846} & \pmb{0.023} / \pmb{0.082} \\
				\bottomrule
			\end{tabular}
		}
	\end{center}
    \vspace{-2mm}
	% \vspace{-8mm}
\end{table}

\noindent\textbf{Effect of Bidirectional Warping in ZoomFI.}
We investigate the respective contributions of forward and backward flow in ZoomFI. The training configurations follow those described in Sec.~\ref{section:5_1}.
As shown in Table~\ref{tab:forward_flow}, using both forward and backward flows consistently improves performance across all evaluation metrics. 
This improvement can be attributed to the complementary sampling properties of the two warping directions.
Specifically, backward warping samples source-frame features at target locations, whereas forward warping maps source locations toward the target. Their complementary properties can improve robustness in regions affected by occlusion and large motion. 
Combining the two warping directions therefore enables more accurate feature alignment and reduces interpolation artifacts.
Visual comparisons in Fig.~\ref{fig:zoomfi} further support this finding: using both flows produces fewer blurring artifacts and structural distortions than using either direction alone.

\section{Conclusion}
In this paper, we introduced the Dual-Camera Smooth Zoom (DCSZ) task to address abrupt transitions in smartphone dual-camera zooming.
To mitigate the lack of practical real-world supervision, we reformulated DCSZ data synthesis as a decoupled geometry–photometry modeling problem and proposed ZoomGS v2, a 3D Gaussian Splatting (3DGS)-based pipeline that synthesizes scalable training data from single-camera datasets by modeling inter-camera geometric and color discrepancies.
Building upon the synthesized data, we developed ZoomFI, a DCSZ-oriented frame interpolation network that leverages bidirectional warping to generate visually plausible intermediate frames.
Extensive experiments on test datasets captured using Xiaomi and Huawei devices demonstrate the data synthesis quality of our approach, and verify that ZoomGS v2 can be extended to various mobile devices.
Future work will explore more accurate camera and ISP modeling as well as efficient deployment on mobile platforms.

\bibliographystyle{splncs04}
\bibliography{egbib}

% \vspace{-10mm}
\clearpage

\ifCLASSOPTIONcaptionsoff
\newpage
\fi

% \vspace{4\baselineskip}
\end{document}